\documentclass[sigconf,nonacm]{acmart}
\AtBeginDocument{%
  }

\setcopyright{acmlicensed}
\copyrightyear{2018}
\acmYear{2018}
\acmDOI{XXXXXXX.XXXXXXX}
\acmConference[Conference acronym 'XX]{Make sure to enter the correct
  conference title from your rights confirmation email}{June 03--05,
  2018}{Woodstock, NY}
\acmISBN{978-1-4503-XXXX-X/2018/06}

\usepackage{tikz}  
\usepackage{subcaption}
\usepackage{algorithm}
\usepackage{algpseudocode}  
\usetikzlibrary{automata, positioning, arrows.meta}
\usepackage{xspace}
\usepackage{enumitem}
\usepackage{tabularx}
\usepackage{flushend}
\usepackage{bm}

\makeatletter

\newcolumntype{Y}{>{\centering\arraybackslash}X}
\begin{document}

\title{SOAR: Real-Time Joint Optimization of Order Allocation and Robot Scheduling in Robotic Mobile Fulfillment Systems}





\settopmatter{authorsperrow=1} 

\author{
  Yibang Tang\textsuperscript{1}, 
  Yifan Yang\textsuperscript{1}, 
  Jingyuan Wang\textsuperscript{1}, 
  Junhua Chen\textsuperscript{1}, 
  Zhen Zhao\textsuperscript{2}
}

\affiliation{%
  \institution{
    \textsuperscript{1} School of Computer Science and Engineering, Beihang University, Beijing, China \\
    \textsuperscript{2} Geekplus Technology Co., Ltd., Beijing, China \\
  }
  \country{China}
}






\renewcommand{\shortauthors}{Tang and Yang, et al.}








\newcommand{\sysname}{\textsc{SOAR}\xspace}
\begin{abstract}
Robotic Mobile Fulfillment Systems (RMFS) rely on mobile robots for automated inventory transportation, coordinating order allocation and robot scheduling to enhance warehousing efficiency. However, optimizing RMFS is challenging due to strict real-time constraints and the strong coupling of multi-phase decisions. Existing methods either decompose the problem into isolated sub-tasks to guarantee responsiveness at the cost of global optimality, or rely on computationally expensive global optimization models that are unsuitable for dynamic industrial environments. 
To bridge this gap, we propose \sysname, a unified Deep Reinforcement Learning framework for real-time joint optimization. \sysname transforms order allocation and robot scheduling into a unified process by utilizing soft order allocations as observations. We formulate this as an Event-Driven Markov Decision Process, enabling the agent to perform simultaneous scheduling in response to asynchronous system events. Technically, we employ a Heterogeneous Graph Transformer to encode the warehouse state and integrate phased domain knowledge. Additionally, we incorporate a reward shaping strategy to address sparse feedback in long-horizon tasks. 
Extensive experiments on synthetic and real-world industrial datasets, in collaboration with Geekplus, demonstrate that \sysname reduces global makespan by 7.5\% and average order completion time by 15.4\% with sub-100ms latency. Furthermore, sim-to-real deployment confirms its practical viability and significant performance gains in production environments. The code is available at \url{https://github.com/200815147/SOAR}.
\end{abstract}





\maketitle

\newcommand{\problem}{DOARS Problem\xspace}

\section{Introduction}
Driven by the explosive demand in global e-commerce, the Robotic Mobile Fulfillment System (RMFS) has emerged as a mainstream solution for modern intelligent warehousing.  RMFS revolutionizes traditional operations by applying a ``goods-to-person'' paradigm. In this system, autonomous robots transport shelves with inventory directly to fixed workstations, eliminating the need for human pickers to traverse the warehouse. This architecture offers significant advantages, including superior throughput efficiency and scalability, making it indispensable for handling the high-volume  order volumes in large-scale logistics centers~\cite{rmfs_survey,rmfs_survey1,rmfs_survey2}.

\begin{figure}[t]
    \centering
    \begin{subfigure}[t]{0.29\linewidth} 
        \centering
        \includegraphics[width=\linewidth, height=4cm, keepaspectratio]{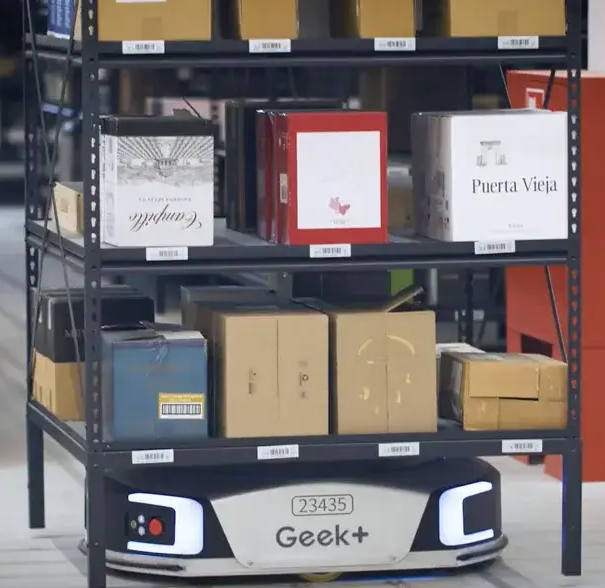} 
        \caption{Robot delivering a shelf}
        \label{fig:intro_top_left}
    \end{subfigure}
    \hfill
    \begin{subfigure}[t]{0.7\linewidth}
        \centering
        \includegraphics[width=\linewidth, height=4cm, keepaspectratio]{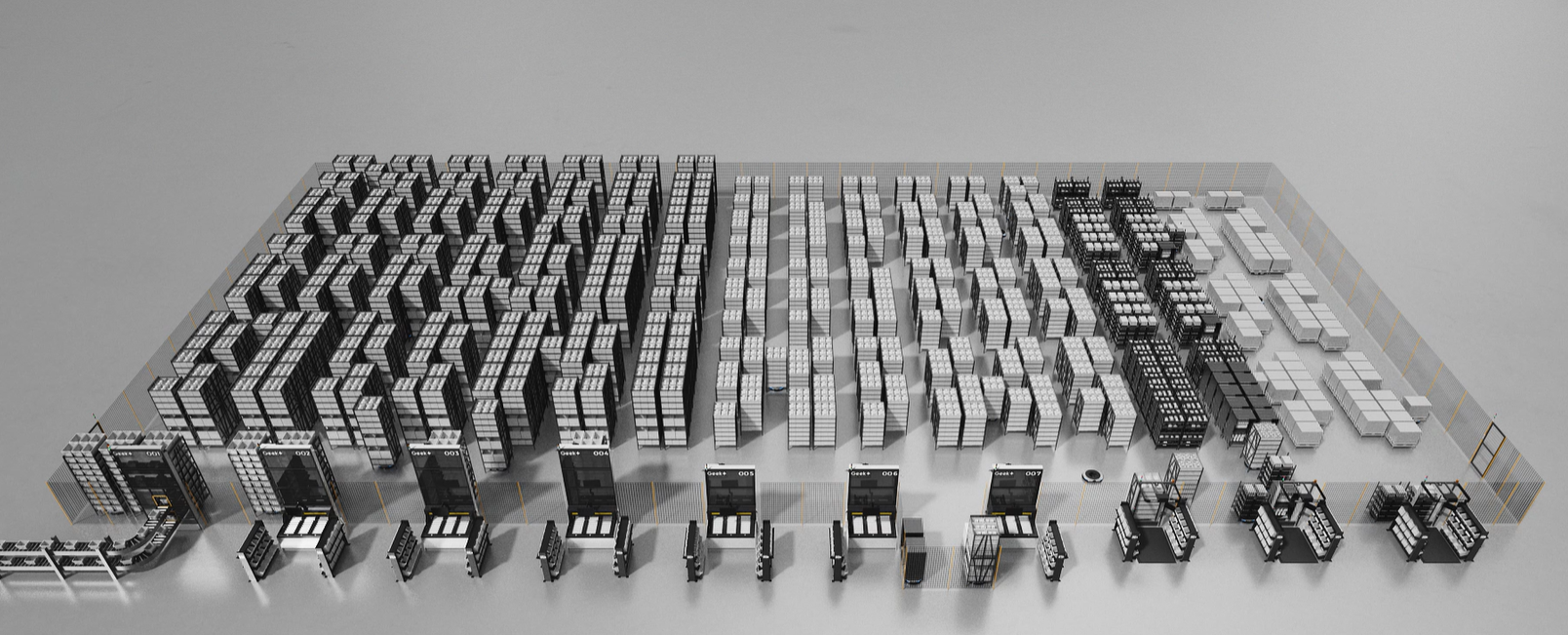} 
        \caption{Top view of the RMFS} 
        \label{fig:intro_top_right} 
    \end{subfigure}

    \begin{subfigure}[b]{1\linewidth} 
        \centering
        \includegraphics[width=1\linewidth]{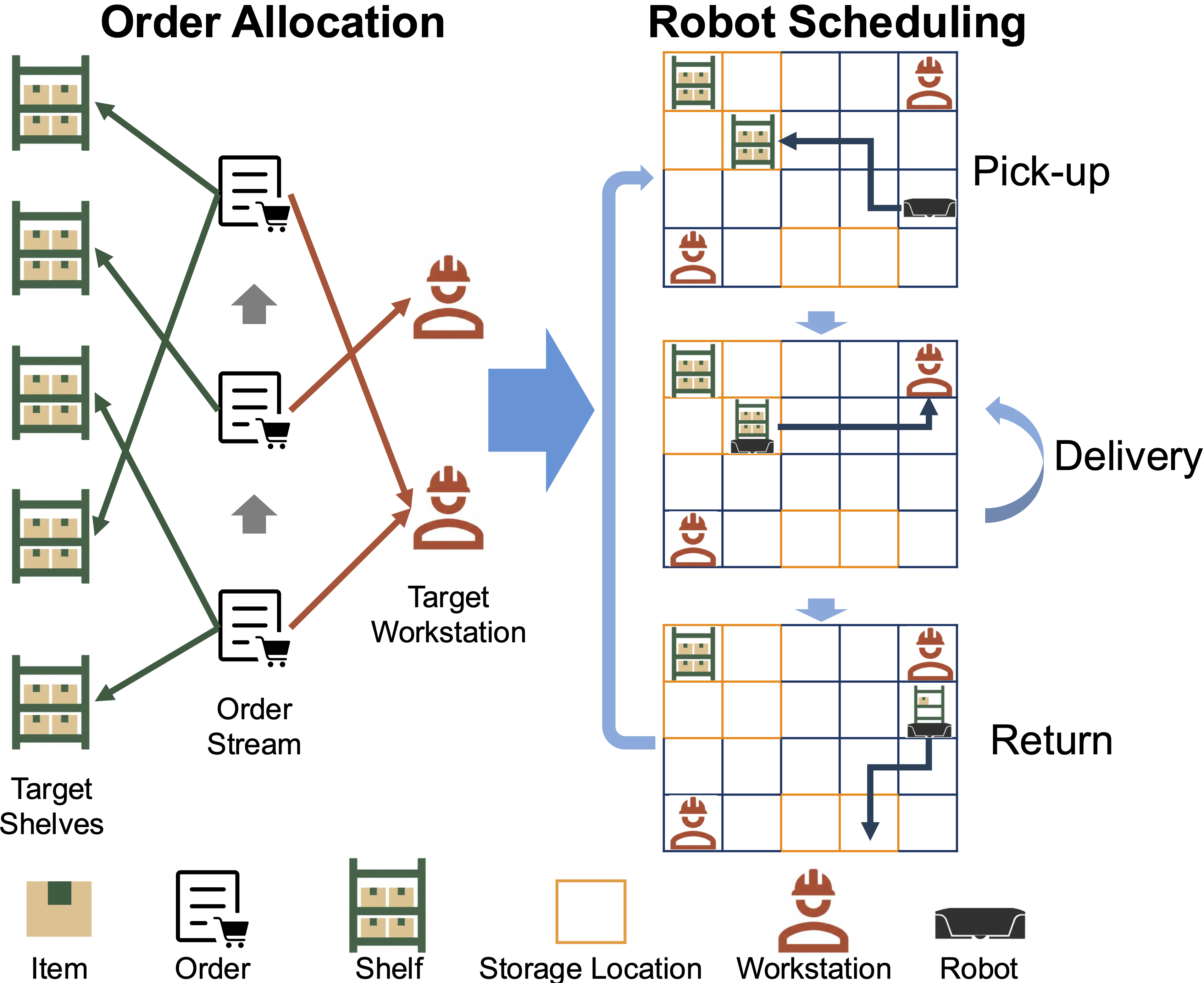} 
        \caption{RMFS workflow}
        \label{fig:intro_bottom}
    \end{subfigure}
    
    \caption{RMFS overview: snapshots and workflow.} 
    \label{fig:intro}
\end{figure}
As illustrated in Figure~\ref{fig:intro}, the RMFS comprises robots for transportation, shelves for items storage, workstations for order fulfillment, and storage locations for shelves placement. Within this environment, the RMFS necessitates a scheduling paradigm comprising two coupled layers that govern a sequential four-phase decision process. Specifically, the \textit{Order Allocation} layer governs the initial order allocation phase, where incoming orders are allocated to specific shelves and workstations. The physical execution is then managed by the \textit{Robot Scheduling} layer, which orchestrates the subsequent three phases: Pick-up to retrieve shelves, Delivery to transport shelves to workstations, and Return to restore shelves to storage locations. Operating at a massive scale with hundreds of robots, the system aims to minimize the total time of order fulfillment. However, achieving this is non-trivial due to intrinsic challenges: \textit{strict real-time constraints} imposed by stochastic order arrivals and rapid robot movements, and the \textit{complex coupling} between multi decision phases.

While existing research has advanced the development of RMFS, current approaches struggle to balance real-time responsiveness and global performance. On the one hand, some research~\cite{wu2025joint} prioritizes real-time responsiveness by adopting a decoupled decision paradigm. These methods decompose the complex system into isolated sub-problems, i.e separate order allocation~\cite{order1,order2,order3} and robot scheduling~\cite{rob1,rob2}. While this decomposition significantly reduces the computational complexity of each individual phase, it inherently restricts the optimization to local objectives and often results in globally suboptimal performance. On the other hand, another targets global optimization by formulating RMFS as a Mixed Integer Programming  model~\cite{teck2022bi,zhao2025learning,yang2021joint} to optimize multiple phases simultaneously. However, these models suffer from computational complexity and are inherently static. They struggle to capture real-time fluctuations efficiently, rendering them unsuitable for large-scale, dynamic industrial applications. Therefore, a unified approach that can simultaneously ensure real-time responsiveness and achieve multi-phase joint optimization is highly desirable.

To bridge this gap, we propose \sysname(\textit{\underline{S}oft \underline{O}rder \underline{A}llocation and \underline{R}obot Scheduling}), a deep reinforcement learning-based framework designed for the \textbf{joint optimization of order allocation and robot scheduling}. Unlike decoupled approaches, our method consolidates them into a unified joint decision-making process. Specifically, we first introduce a \textit{Soft Order Allocation} mechanism. Instead of executing immediate fixed allocation, this module dynamically calculates matching degree between orders, shelves, and workstations. These matching degrees serve as inputs for the subsequent phase, effectively fusing the previously isolated phases into a unified optimization process. Building upon this, we formulate the joint decision-making process as an Event-Driven Markov Decision Process. The operational cycle is driven asynchronously by distinct events. At each decision step, the system gets the real-time global state with the soft allocation information to generate actions, which drive the continuous evolution of the system environment.


In collaboration with Geekplus\footnote{\url{https://www.geekplus.com/en}}, a leading warehouse robotics company, we evaluated \sysname on both large-scale synthetic and real-world industrial datasets. Experimental results demonstrate that our framework significantly outperforms state-of-the-art solutions, reducing the global \textit{makespan} by 7.5\% and average order completion time by 15.4\%. 
Furthermore, we successfully executed a sim-to-real deployment in a physical production environment. Extensive on-site testing confirms that \sysname yields consistent and significant improvements across all key metrics under real-world conditions. 
Crucially, these performance gains are achieved while maintaining sub-100ms response latency. \sysname successfully bridges the gap between real-time responsiveness and global performance, exhibiting immense potential for broad industrial application.

The main contributions of this paper are summarized as follows:
\begin{itemize}[leftmargin=*]
    \item 
    To the best of our knowledge, \sysname is one of the \textbf{initial attempts into the dynamic joint optimization} of order allocation and robot scheduling in RMFS. 
    \item 
    We propose a novel decision paradigm that synergizes \textbf{soft order allocation} with an \textbf{event-driven markov decision process}. This paradigm effectively dissolves the boundary between order allocation and robot scheduling, transforming them into a \textbf{unified optimization process}. Building upon this, we develop a reinforcement learning-based system that presents a new solution for addressing complex scheduling challenges in RMFS.
    \item 
    Extensive experiments on both large-scale synthetic and real-world industrial datasets demonstrate that \sysname significantly outperforms state-of-the-art solutions, reducing the global makespan by \textbf{7.5\%} and average order completion time by \textbf{15.4\%} while maintaining \textbf{sub-100ms latency}. Furthermore, we successfully bridged the reality gap through a \textbf{sim-to-real deployment} in production environments, validating that \sysname yields significant improvements under real-world constraints.
\end{itemize}
\section{Preliminaries and System Overview}
\subsection{Preliminaries}
The warehouse environment is modeled as a 2D grid map of size $H \times W$. It contains following entities:

\begin{definition}[Item]
The basic unit of inventory, representing a product category. The warehouse manages $N_k$ unique item types.
\end{definition}


\begin{definition}[Order]
    An order $o = (t_o, \bm{d}_o)$ consists of an arrival time $t_o$ and a demand list $\bm{d}_{o} \in \mathbb{Z}^{N_k}$, where $\bm{d}_o[i]$ denotes the demanded quantity of the $i$-th item. $\mathcal{O}$ denotes the set of orders.
\end{definition}

\begin{definition}[Shelf]
    A shelf $s=(x_s,y_s,\bm{q}_s)$ consists of its position $(x_s,y_s)$ and inventory $\bm{q}_{s} \in \mathbb{Z}^{N_k}$, where $\bm{q}_s[i]$ is the quantity of the $i$-th item on shelf. $\mathcal{S}$ denotes the set of shelves with size $N_s$.
\end{definition}

\begin{definition}[Storage Location]
    A storage location $l$ is defined by a tuple $(x_{l}, y_l)$, where $(x_l,y_l)$ denotes its fixed position for holding shelves. $\mathcal{L}$ denotes the set of storage locations with size $N_l$.
\end{definition}

\begin{definition}[Workstation]
    A workstation $w$ is defined by a tuple $(x_w, y_w)$, where $(x_w,y_w)$ denotes its fixed position. Pickers at workstations retrieve items from shelves to fulfill order demands. $\mathcal{W}$ denotes the set of workstations with size $N_w$.
\end{definition}

\begin{definition}[Robot]
    A robot $r$ is defined by a tuple $(x_{r}, y_r, s_{r})$, where $(x_r,y_r)$ denotes the position, $s_{r} \in \mathcal{S} \cup \{\emptyset\}$ indicates the loaded shelf (or $\emptyset$ if unloaded). Unloaded robots can traverse storage locations by passing under the shelves. The robot transports the shelf to the workstation. $\mathcal{R}$ denotes the set of robots with size $N_r$.
\end{definition}

\begin{definition}[\underline{D}ynamic \underline{O}rder \underline{A}llocation and \underline{R}obot \underline{S}cheduling \underline{P}roblem, \problem]
    Given the warehouse and dynamically arriving orders $\mathcal{O}$, the \problem involves making decisions on \textbf{1) Order Allocation}: Allocating each order to a workstation and shelves that contain all required items; \textbf{2) Robot Scheduling}: Scheduling robots to specific targets, such as moving to a storage location with shelf for pick-up, a workstation for delivery or an empty storage location for return shelf.
    The goal is to minimize the \textit{makespan} $M$:
    \begin{equation}
        \min \quad M = \max_{r \in \mathcal{R}} T_r,
    \end{equation}
    where $T_r$ denotes the elapsed time for robot $r$ from the start to the completion of the scheduling process.
\end{definition}

\subsection{System Overview}
\begin{figure}[t]
    \centering
    \includegraphics[width=1.05\linewidth]{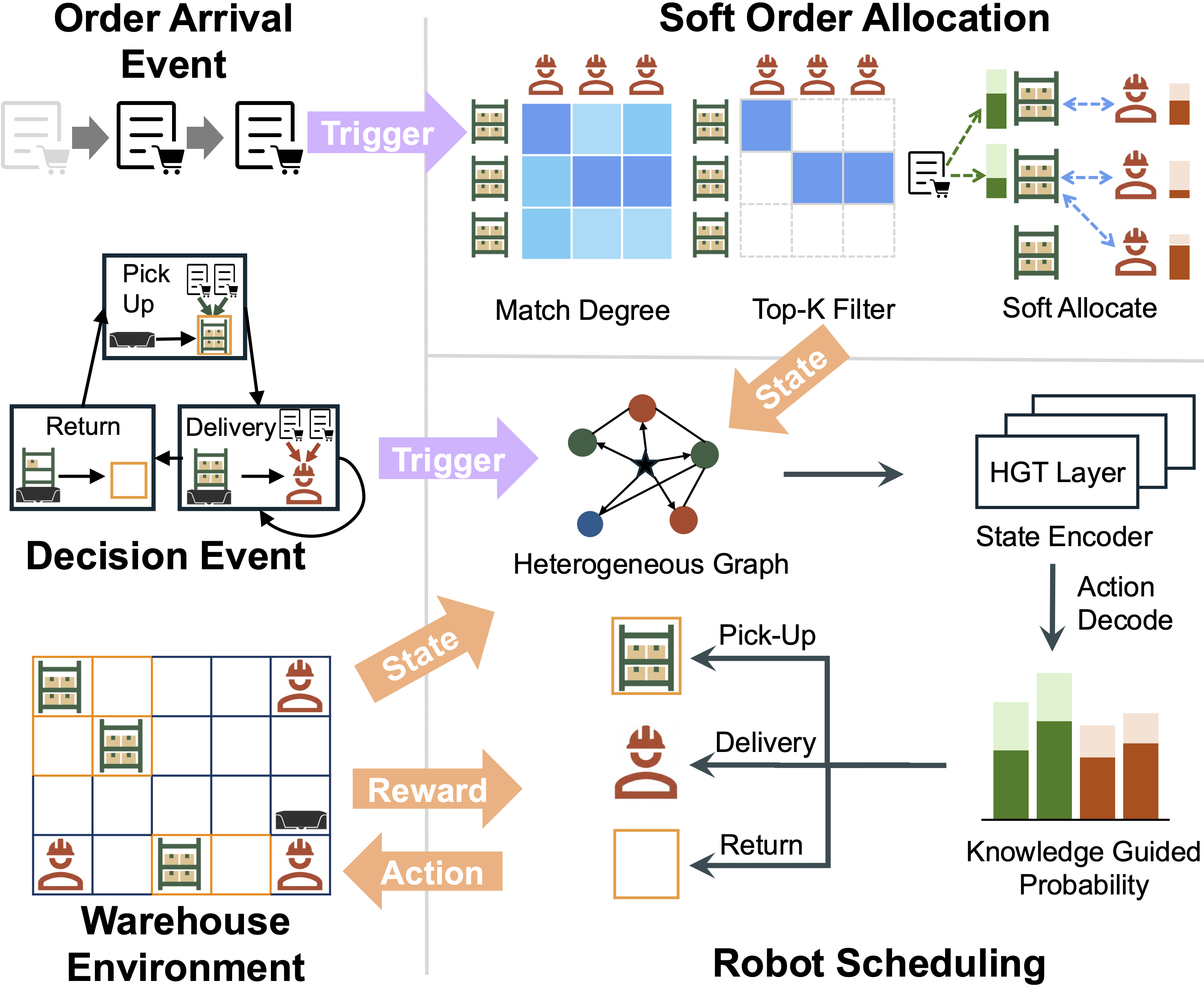}
    \caption{The overall framework of \sysname. Purple arrows indicate that an event triggered a module, and orange arrows indicate the information flow in an Event-Driven MDP.}
    \label{fig:main_framework}
\end{figure}

The framework of \sysname is illustrated in Figure~\ref{fig:main_framework}, comprising the warehouse environment, a \textit{Soft Order Allocation} module, and a \textit{Robot Scheduling} module. The environment operates in an event-driven manner, governed by two key events: order arrivals and decision events. Upon order arrival, \textit{Soft Order Allocation} evaluates allocation candidates to provide preliminary guidance, deferring the final commitment to the subsequent robot scheduling phase. Triggered by decision events, \textit{Robot Scheduling} determines the robot's next destination based on prior soft allocations and the current state. This module outputs an action while finalizing specific order allocations, driving state transitions that generate new events.

\section{Soft Order Allocation}
\label{sec:soft_allocation}

Traditional methods that immediately allocate orders to fixed shelves and workstations overlook the system dynamics that evolve between allocation and the transportation of shelves to workstations. \sysname employs a \textit{soft allocation} mechanism. This process is triggered instantaneously upon the arrival of each new order. Its objective to update: 1) \textit{Soft Orders Set} $\mathcal{O}_s$, represents the orders potentially allocated to the shelf $s$. The collection of soft order sets across all shelves is denoted by $\mathcal{O}_{(\cdot)}$.
2) \textit{Heat Vectors} $\bm{h}^{S} \in \mathbb{R}^{N_s},\bm{h}^{W} \in \mathbb{R}^{N_w}$, represent the potential of shelves and workstations for allocation to the orders that have already arrived.  These subsequently serves as guidance for robot scheduling. For every incoming order $o$, the following steps are executed sequentially:
\paragraph{Step 1: Construction of Matching Degree Matrix}
For the incoming order $o$, we construct a matching degree matrix $\bm{V}^{(o)} \in \mathbb{R}^{N_s \times N_w}$, where $\bm{V}^{(o)}_{s,w}$ quantifies the suitability of delivering shelf $s$ to workstation $w$ to fulfill order $o$. A higher $\bm{V}^{(o)}_{s,w}$ indicates that shelf $s$ is highly relevant to order $o$ and is close to workstation $w$. Formally:
\begin{equation}
    \bm{V}^{(o)}_{s,w} = \frac{\sum_{i=1}^{N_k} \min(\bm{d}_o[i], \bm{q}_s[i])}{\text{dist}(s, w) + \epsilon},
    \label{eq:matching_degree}
\end{equation}
where $\text{dist}(s,w)$ denotes the distance between the shelf and the workstation, and $\epsilon$ is a small constant to prevent division by zero. The numerator represents the quantity of items retrievable from the shelf to satisfy the order, while the denominator denotes the distance required to transport the shelf to the workstation.

\paragraph{Step 2: Top-$K$ Candidate Shelves Filtering}
In this step, we filter out the most promising candidates shelves to avoid the heavy computational burden arising from the large shelf number.

For each workstation $w$, we select the subset of shelves $\mathcal{C}^{(o)}_w$ with the Top-$K$ highest matching degrees, where $K$ represents the size of the candidate set. The candidate set for workstation $w$ is defined as:
\begin{equation}
    \mathcal{C}^{(o)}_w = \left\{ s\in \mathcal{S} \mid \text{rank}(\bm{V}^{(o)}_{s,w}) \leq K \right\},
    \label{eq:filtering}
\end{equation}
where $\text{rank}(\bm{V}^{(o)}_{s,w})$ corresponds to the position of $\bm{V}^{(o)}_{s,w}$ within $\bm{V}^{(o)}_{:,w}$ (column $w$ of $\bm{V}^{(o)}$) arranged in descending order. This step prunes the soft allocation space, retaining only the Top-$K$ shelves that are most likely to be selected for order $o$ at workstation $w$.



\paragraph{Step 3: Soft Allocation and Heat Vectors Update} This step involves two parallel processes: maintaining the \textit{Soft Orders Set} $\mathcal{O}_s$ and accumulating values for the \textit{Heat Vectors} $\bm{h}^{S} \in \mathbb{R}^{N_s},\bm{h}^{W} \in \mathbb{R}^{N_w}$of both shelves and workstations.

First, let $\mathcal{C}^{(o)}_{all} = \bigcup_{w \in \mathcal{W}} \mathcal{C}^{(o)}_w$ be the union of all candidate shelves identified in Step 2. The updates are defined as follows:
\begin{itemize}[leftmargin=*]
\item \textbf{Soft Orders Set Update ($\mathcal{O}_{(\cdot)}$):} For every shelf $s$ that appears in the candidate list of any workstation, we consider order $o$ is potentially allocated to the shelf. We update the set $\mathcal{O}_s$ for each shelf $s \in \mathcal{C}^{(o)}_{all}$ by appending the current order:
\begin{equation}
    \mathcal{O}_s \leftarrow \mathcal{O}_s \cup \{o\}.\ \ \ (s\in \mathcal{C}^{(o)}_{all})
    \label{eq:update_Os}
\end{equation}
This set $\mathcal{O}_s$ records which orders are competing for shelf $s$, providing semantic information for the subsequent robot scheduling. Notably, an order may be allocated to at most $N_w\times K$ shelves. 


\item \textbf{Shelf Heat Update ($\bm{h}^{S}$):} For each shelf $s \in \mathcal{C}^{(o)}_{all}$, we accumulate the matching degrees from all workstations that selected it:
\begin{equation}
    h^S_s \leftarrow h^S_s + \sum_{w \in \mathcal{W}} \mathbb{I}(s \in \mathcal{C}^{(o)}_w) \cdot \bm{V}^{(o)}_{s,w},
    \label{eq:update_hs}
\end{equation}
where $\mathbb{I}(\cdot)$ is the indicator function. The update of $\bm{h}^{S}$ guides robots to retrieve shelves that are urgently needed. 

\item \textbf{Workstation Heat Update ($\bm{h}^{W}$):}  For each workstation $w \in \mathcal{W}$, we aggregate the degrees of its top-$K$ candidates:
\begin{equation}
    h^W_w \leftarrow h^W_w + \sum_{s \in \mathcal{C}^{(o)}_w} \bm{V}^{(o)}_{s,w}.
    \label{eq:update_hw}
\end{equation}
The update of $\bm{h}^{W}$ indicates which workstations have a better supply of required goods nearby.
\end{itemize}

$\bm{h}^W,\bm{h}^S,\mathcal{O}_{(\cdot)}$ form the global context that bridges the gap between order allocation and robot scheduling. In the subsequent robot scheduling phase, the policy model utilizes these information with the real-time status of robots as state inputs and the precise allocation is finalized.

\section{Robot Scheduling and Final Order Allocation}
\label{sec:unified_model}
The scheduling module monitors the dynamic events in warehouse to schedule robots and finalize orders allocation.
To achieve real-time response to asynchronous events, we formulate the scheduling process as an \textbf{Event-Driven Markov Decision Process (ED-MDP)}. Building upon the ED-MDP, we implement a scheduling module based on Deep Reinforcement Learning (DRL). By modeling the warehouse environment as a heterogeneous graph, this model utilizes an HGT module to encode environment states and incorporates Phase-Knowledge to make real-time decisions.

\subsection{Event-Driven Markov Decision Process}
\label{subsec:mdp_framework}
\begin{figure}[t]
    \centering
    \begin{tikzpicture}[
        ->, >=Stealth, 
        shorten >=1pt, 
        semithick,
        auto,
        tiny_state/.style={
            circle, 
            draw, 
            align=center,       
            inner sep=1pt,      
            minimum size=1.3cm, 
            font=\small    
        },
        every edge quotes/.style = {font=\tiny, inner sep=1pt} 
    ]
        \node[tiny_state, initial, initial text=] (s0) {Idle};
        \node[tiny_state] (s1) [right=1.1cm of s0] {Pick-up\\Completion}; 
        \node[tiny_state] (s2) [right=1.1cm of s1] {Delivery\\Completion};
        \path
            (s2) edge [loop above, min distance=5mm, in=70, out=110] node {Delivery} (s2)
            (s0) edge [bend left=0]  node {Pick-up} (s1)
            (s1) edge [bend left=0]  node {Delivery} (s2)
            (s2) edge [bend left=35]  node [below, yshift=1pt] {Return} (s0);
    \end{tikzpicture}
    \caption{The Cycle of Event Generation and Policy Actions.}
    \label{fig:trans}
\end{figure}
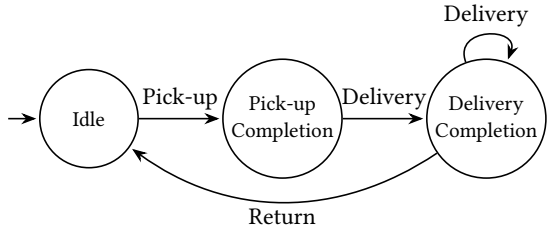

The scheduling cycle is governed by three distinct types of events, each necessitating a event-specific action:
\begin{itemize}[leftmargin=*]
    \item \textbf{Idle:} Triggered when status of any robot becomes \textsc{IDLE}. The robot selects a storage location with shelf to pick up. Action space $\mathcal{A}=\{l\in\mathcal{L}\ |\ l\ \text{is occupied by a shelf}\}$.
    \item \textbf{Pick-up Completion:} Triggered when a robot picks up a shelf. The robot selects a target workstation to deliver the carried shelf. Action space $\mathcal{A}=\mathcal{W}$.
    \item \textbf{Delivery Completion:} Triggered when a picker at a workstation completes the picking of items from a shelf transported by a robot. The robot selects next workstation to delivery or selects an empty storage location to restore the shelf. After restoring the shelves, the robot status becomes \textsc{IDLE} and completes one cycle of scheduling. Action space $\mathcal{A}=\mathcal{W}\cup\{l\in\mathcal{L}\ |\ l\ \text{is empty}\}$.
\end{itemize}
The actions taken by the policy model trigger the generation of further events, as illustrated in Figure~\ref{fig:trans}.


\subsection{State Representation}
\label{subsec:state_encoding}
To make informed decisions, the agent requires a comprehensive understanding of the warehouse environment state. Given the heterogeneity of the warehouse entities, we formulate them as a heterogeneous graph and employ a Heterogeneous Graph Transformer (HGT) as the state encoder to process the complex global state.

At any decision step $t$ triggered by an event, the raw state $s_t$ serves as the input, composed of two distinct states:
\begin{itemize}[leftmargin=*]
    \item \textbf{Entity State ($s^{ent}_t$):} Represents the physical reality of the warehouse, including the spatial coordinates of entities, robot availability, and real-time storage occupancy.
    \item \textbf{Soft Allocation State ($s^{soft}_t$):} Represents the guidance from soft order allocation, specifically heat vectors $\bm{h}^{S}$, $\bm{h}^{W}$ and the soft orders sets $\mathcal{O}_{(\cdot)}$.
\end{itemize}
Formally, the raw state $s_t$ at step $t$ is constructed by concatenating these two feature groups: $s_t = [s^{ent}_t; s^{soft}_t]$.

\textit{Entities Selection and Feature Projection.}
We choose robots, workstations and storage locations as the encoded entities, where shelves are considered as features of robots or storage locations rather than independent entities. For each encoded entity $v$, we extract its corresponding raw attributes $s_t[v]$ and project them into a dense vector $\bm{h}^0_v$ via type-specific MLP. The details of projection is provided in Appendix~\ref{app:encode}.

\textit{Graph Construction and HGT Encoding.}
We construct a heterogeneous graph $\mathcal{G}_t = (\mathcal{V}_t, \mathcal{E}_t)$, where $\mathcal{V}_t$ consists of all encoded entities and the special \textit{Phase Node} with the event type specific learnable embedding $\bm{h}^0_{phase}$ to model the impact of different events on decision-making. And $\mathcal{E}_t$ consists of bi-directional edges between all heterogeneous entity pairs weighted by distance, and directed edges from \textit{Phase Node} to other nodes.

We employ a stack of $L$ HGT layers to perform global message passing on the constructed graph. The representation of the $\ell$-th layer is denoted as $\bm{h}_i^{(\ell)}$. For each target node $i$, it is updated by aggregating messages from neighbors under specific relations:
\begin{equation}
\bm{h}_i^{(\ell+1)} = \text{HGT}\left( \bm{h}_i^{(\ell)}, \left\{ \bm{h}_j^{(\ell)} \right\} \right), j \in \mathcal{N}(i),\quad \ell=0,\dots,L-1,
\end{equation}
where HGT operator applies relation-specific Graph Attention mechanisms~\cite{gatv2}. The implementation details are provided in Appendix~\ref{app:hgt}.

\subsection{Action Generation}
\label{subsec:action_transition}
Once the state is encoded, the \textbf{Phase-Knowledge-Guided Decoder} interprets these representations to generate optimal actions, which subsequently drive the system state transition.

The decoder fuses the state embedding with phase-specific priors to generate actions. For an active robot $r^*$ and any candidate target $i$, the logit $z_i$ is computed via a MLP augmented by a \textit{phase-specific bias}:
\begin{equation}
    z_i = \text{MLP}_{\text{dec}} ( \bm{h}_i^{(L)} \Vert \bm{h}_{r^*}^{(L)} ) + b_i
\end{equation}
where $\bm{h}_i^{(L)}$ and $\bm{h}_{r^*}^{(L)}$ are the final-layer representations from the HGT encoder. The bias $b_i$ injects phase-specific priors:
\begin{equation}
    b_i = 
    \begin{cases} 
      \log(h^{S}_i + \epsilon) & \text{phase: \textsc{Pick-up}} \\
      -\log(u_i + \epsilon) & \text{phase: \textsc{Delivery}} \\
      -\log(\text{dist}(r^*, i) + \epsilon) & \text{phase: \textsc{Return}}
   \end{cases}
\end{equation}
The rationale behind these priors is three-fold: 
(i) \textbf{Pick-up}: Prioritizing shelves with higher heat $h^{S}_i$ to maximize fulfillment potential; 
(ii) \textbf{Delivery}: $u_i$ denotes the workload of $w_i$ (The quantity of unprocessed items). Penalizing workstations with lower $u_i$ to ensure global load balancing and reduce queuing; 
(iii) \textbf{Return}: Minimizing travel distance $\text{dist}(r^*, i)$ to accelerate robot turnover. 

Then we apply a validity mask $\bm{m}$ according to the action space of current event to yield the policy distribution: 
\begin{equation}
\pi(a_t = i | s_t) = \text{Softmax}(z_i + m_i).
\end{equation}
Finally, we sample from this distribution to obtain action $a_t$.

\subsection{Final Order Allocation}

Simultaneously with the robot's decision $a_t$, the system converts the soft order allocations into deterministic allocations to fulfill order requirements. Specifically, the allocation of orders to shelves and the update of soft information is executed during the pick-up phase, while the assignment of orders to workstations is performed during the delivery phase.

\textit{Pick-up (Order-to-Shelf Allocation).}
During the soft allocation phase, the items on the shelf $s$ may not be able to meet all demands of order in soft orders set $\mathcal{O}_s$ (Formula~\ref{eq:update_Os}). We need to determine which orders can be fulfilled and which cannot, and simultaneously complete the step of allocating the fulfillable orders to the shelves.
Let the robot pick up shelf $s^*$. We define two disjoint sets: $\mathcal{O}_{feas}$ for orders that can be fully satisfied by the current shelf items $\bm{q}_{s^*}$, and $\mathcal{O}_{rem}$ for orders that remain unsatisfied. 
Specifically, we iterate through each order $o \in \mathcal{O}_{s^*}$. If the current items suffice (i.e., $\bm{q}_{s^*}[i] \ge \bm{d}_o[i],\forall 1\le i\le N_k$), we allocate $o$ to $s^*$:
\begin{equation}
    \mathcal{O}_{feas}\gets\mathcal{O}_{feas}\cup \{o\},
\end{equation}
 and subtract order demand from shelves' items:
\begin{equation}
    \bm{q}_{s^*} \leftarrow \bm{q}_{s^*} - \bm{d}_o.
\end{equation}
Otherwise, we add $o$ to $\mathcal{O}_{rem}$: 
\begin{equation}
    \mathcal{O}_{rem}\gets\mathcal{O}_{rem} \cup \{o\},
\end{equation}
for subsequent processing.

\textit{Update Soft Information.} Given that orders in $\mathcal{O}_{s^*}$ are possibly soft allocated to multiple shelves, it is necessary to eliminate the influence of them on other shelves and workstations. Specifically, for every $o \in \mathcal{O}_{s^*}$, we remove it from all soft orders sets: 
\begin{equation}
    \mathcal{O}_s \leftarrow \mathcal{O}_s \setminus \{o\},\quad\forall s\in \mathcal{S}
    \label{eq:update_Os_2}
\end{equation}
and subtract its contribution to the heat vectors for shelves and workstations $\bm{h}^{S}, \bm{h}^{W}$:
\begin{equation}
    h^S_s \leftarrow h^S_s - \sum_{w \in \mathcal{W}} \mathbb{I}(s \in \mathcal{C}^{(o)}_w) \cdot \bm{V}^{(o)}_{s,w},\quad\forall s\in \mathcal{S}
    \label{eq:update_hs1}
\end{equation}
\begin{equation}
    h^W_w \leftarrow h^W_w - \sum_{s \in \mathcal{C}^{(o)}_w} \bm{V}^{(o)}_{s,w}.\quad\forall w\in \mathcal{W}
    \label{eq:update_hw1}
\end{equation}

\textit{Delivery (Order-to-Workstation Allocation and Remainder Handling).} Finally, we need to complete order-to-workstation allocation and selecting shelves for orders in $\mathcal{O}_{rem}$. Let robot deliver shelf $s^*$ to workstation $w^*$. Orders in $\mathcal{O}_{feas}$ enter the workstation's queue and are fulfilled after the picker completes the picking operation. 
For every order in $\mathcal{O}_{rem}$, we rank remaining shelves $s\in \mathcal{S}\setminus \{s^*\}$ based on their matching degree with $w^*$ (i.e., $\bm{V}^{(o^*)}_{s,w*}$) and greedily select them in descending order until all requirements are met. Further implementation details are available in the Appendix~\ref{app:hard_allocate}, and analysis of its impact on performance are available in the Appendix~\ref{app:long_tail}.

\section{Training Paradigm}
\label{sec:training}

We employ a deep reinforcement learning approach to optimize the event-driven markov decision process, where the objective of the RMFS is to minimize the \textit{makespan}. However, directly optimizing this objective presents significant challenges: (1) \textit{Sparse Feedback:} The makespan is only determined at the end of the episode, causing severe delayed credit assignment; \textit{(2) Zero Gradient Feedback:} The non-differentiable nature of the makespan yields zero gradients for non-bottleneck robots, preventing them from learning to improve efficiency. To address these issues, we propose a dense reward mechanism based on \textbf{$p$-norm Reward Shaping} serving as a smooth surrogate of the original makespan objective.

\subsection{$p$-norm Reward Shaping}
We address the sparse and non-smooth nature of the objective by approximating the $L_\infty$ landscape with a smoother $L_p$ surrogate. Let $\bm{T}_t \in \mathbb{R}^{N_r}$ denote the vector of accumulated active times for all robots at decision step $t$. We define a state potential function $\Phi(s_t)$ based on the scaled $L_p$-norm:
\begin{equation}
    \Phi(s_t) = \left( \frac{1}{N_r}\sum_{r \in \mathcal{R}} ( T_{r, t} )^p \right)^{1/p}
\end{equation}
where $p$ is a hyperparameter controlling sensitivity to outliers. Following the logic of Potential-Based Reward Shaping (PBRS)~\cite{pbrs}, we construct a dense reward $r_t^{\text{shape}}$ for the transition $s_t \to s_{t+1}$:
\begin{equation}
    r_t^{\text{shape}} = - \left( \gamma \Phi(s_{t+1}) - \Phi(s_t) \right),
\end{equation}
where $\gamma$ is the discount rate. By minimizing the incremental growth of the $p$-norm, the agent receives immediate feedback for action.

\subsection{Time-Aware PPO Optimization}
\label{subsec:optimization}

We employ PPO~\cite{ppo} to optimize the policy. A critical challenge in our Event-Driven MDP is the \textit{variable time intervals} between decision steps. We address this by incorporating the physical time duration into the discount mechanism. Let $\tau_t$ denote the physical timestamp of the $t$-th decision event. The duration of the transition $s_t \to s_{t+1}$ is given by $\Delta \tau_t = \tau_{t+1} - \tau_t$. We define the dynamic discount factor as $\gamma_t = \gamma^{\Delta \tau_t}$, where $\gamma$ is the standard discount rate per unit time.

To reduce variance, we utilize Generalized Advantage Estimation (GAE)~\cite{gae}. The time-aware TD-error $\delta_t$ and the advantage $\hat{A}_t$ are computed as:
\begin{equation}
    \delta_t = r_t^{\text{shape}} + \gamma^{\Delta \tau_t} V(s_{t+1}) - V(s_t),
\end{equation}
\begin{equation}
    \hat{A}_t = \sum_{k=0}^{\infty} (\lambda \gamma)^{T_{k}} \delta_{t+k} \quad \text{where } T_k = \sum_{j=0}^{k-1} \Delta \tau_{t+j}
\end{equation}
Here, we apply the discount cumulatively over physical time. The final training objective is to maximize the following function:
\begin{equation}
    J(\theta) = \mathbb{E}_t \left[ \mathcal{L}^{\text{CLIP}}_t(\theta) - c_1 \mathcal{L}^{\text{VF}}_t(\theta) + c_2 S[\pi_\theta](s_t) \right]
\end{equation}
where $\mathcal{L}^{\text{CLIP}}$ is the standard PPO surrogate objective, $S$ denotes the entropy bonus to encourage exploration, and the value function loss is defined as $\mathcal{L}^{\text{VF}}_t = \| V(s_t) - (r_t^{\text{shape}} + \gamma^{\Delta \tau_t} V(s_{t+1})) \|^2$. The raw features of value function is detailed in Appendix~\ref{app:encode}.
\begin{table*}[t]
  \centering
  \caption{Performance on synthetic and real-world datasets.}
  \renewcommand{\arraystretch}{0.85} 
    \begin{tabular}{l|ccc|ccc|ccc}
    \toprule[1.1pt]
    \textbf{Dataset} & \multicolumn{3}{c|}{\textbf{Synth-Small}} & \multicolumn{3}{c|}{\textbf{Synth-Medium}} & \multicolumn{3}{c}{\textbf{Synth-Large}} \\
    \midrule
    \textbf{Metrics} & \multicolumn{1}{c}{Obj ($\downarrow$)} & CompT ($\downarrow$) & \multicolumn{1}{c|}{Time ($\downarrow$)} & Obj ($\downarrow$) & CompT ($\downarrow$) & 
    \multicolumn{1}{c|}{Time ($\downarrow$)} & Obj ($\downarrow$) & CompT ($\downarrow$) & Time ($\downarrow$) \\
    \midrule
    
    SQF+Nearest & 1669.47 & 716.06 & 0.81s & 3079.12 & 1208.14 & 1.58s & 4760.42 & 1643.87 & 2.50s \\
    SQF+Earliest & 1856.94 & 776.67 & 0.79s & 3158.61 & 1213.41 & 1.53s & 4617.98 & 1633.91 & 2.68s \\
    SQF+TSP & 1375.56 & 606.47 & 1.05s & 2347.68 & 916.51 & 1.91s & 3684.72 & 1261.73 & 3.61s \\
    SQF+PSMDRL & 1278.03 & 502.89 & 4.09s & 2170.25 & 718.67 & 10.02s & 3374.01 & 947.52 & 16.72s \\
    
    WLB+Nearest & 1231.12 & 474.12 & 1.07s & 2010.73 & 648.02 & 1.97s & 2918.21 & 816.41 & 3.62s \\
    WLB+Earliest & 1749.07 & 757.01 & 1.09s & 2639.03 & 1002.58 & 2.26s & 3510.60 & 1140.82 & 4.26s \\
    WLB+TSP & 1187.91 & 467.30 & 1.40s & 1878.96 & 654.27 & 3.57s & 2761.08 & 852.32 & 7.88s \\
    WLB+PSMDRL & 1316.54 & 615.60 & 4.27s & 2005.14 & 635.90 & 9.19s & 2856.74 & 827.10 & 17.02s \\
    
    OR Tools+Nearest & 1173.21 & 539.01 & 4m 12s & 2160.55 & 868.56 & 10m 21s & 3686.29 &  904.32 & 17m 31s \\
    OR Tools+Earliest & 1137.10 & 577.78 & 4m 7s & \underline{1629.95} & 695.51 & 9m 8s & 3004.79 & 780.09 & 15m 35s \\
    OR Tools+TSP & \underline{998.10} & 442.57 & 4m 8s & 1635.95 & \underline{420.24} & 9m 46s & \underline{2446.59} & \underline{547.90} & 15m 45s \\
    \midrule
    
    JOTP & 1272.98 & 501.99 & 1.74s & 2138.72 & 703.33 & 3.97s & 3124.54 & 829.30 & 8.01s \\
    SABS&1023.84&\underline{347.24}&4m 10s&1925.40&560.38&10m 54s&2614.15&705.17&20m 9s\\
    \midrule
    
    \sysname & \textbf{811.01} & \textbf{243.25} & 8.74s & \textbf{1387.82} & \textbf{289.94} & 20.43s & \textbf{2160.47} & \textbf{362.08} & 39.60s \\
    Improvement & 18.7\% & 29.9\% &-& 14.9\% & 31.0\% &-& 11.7\% & 33.9\% &-\\
    \midrule
    \midrule
    \textbf{Dataset} & \multicolumn{3}{c|}{\textbf{Real-Small}} & \multicolumn{3}{c|}{\textbf{Real-Medium}} & \multicolumn{3}{c}{\textbf{Real-Large}} \\
    \midrule
    \textbf{Metrics} & \multicolumn{1}{c}{Obj ($\downarrow$)} & CompT ($\downarrow$) & \multicolumn{1}{c|}{Time ($\downarrow$)} & Obj ($\downarrow$) & CompT ($\downarrow$) & \multicolumn{1}{c|}{Time ($\downarrow$)} & Obj ($\downarrow$) & CompT ($\downarrow$) & Time ($\downarrow$) \\
    \midrule

    SQF+Nearest & 618.62 & 259.91 & 0.60s & 966.22 & 435.03 & 1.01s & 1445.08 & 611.81 & 1.64s \\
    SQF+Earliest & 604.61 & 242.27 & 0.60s & 993.60 & 424.71 & 0.99s & 1585.79 & 675.22 & 1.62s \\
    SQF+TSP & 614.88 & 242.29 & 0.62s & 955.79 & 404.59 & 1.32s & 1533.20 & 649.67 & 1.90s \\
    SQF+PSMDRL & 550.59 & 218.33 & 2.50s & 901.60 & 365.28 & 5.68s & 1265.79 & 580.30 & 8.19s \\
    WLB+Nearest & 494.19 & 201.41 & 0.63s & 743.77 & \underline{258.55} & 1.44s & \underline{916.77} & \underline{315.23} & 2.07s \\
    WLB+Earliest & 606.33 & 257.18 & 0.73s & 892.97 & 330.77 & 1.29s & 1377.31 & 587.61 & 2.30s \\
    WLB+TSP & 520.75 & 204.66 & 0.89s & 770.33 & 275.02 & 6.57s & 940.60 & 348.00 & 1m 29s \\
    WLB+PSMDRL & 480.92 & 222.28 & 2.40s & 811.50 & 319.33 & 4.15s & 926.90 & 347.26 & 6.28s \\
    
    OR Tools+Nearest & \underline{414.06} & \underline{184.50} & 1m 34s & 761.74 & 281.98 & 3m 58s & 1229.56 & 447.28 & 7m 36s \\
    OR Tools+Earliest & 449.13 & 204.98 & 1m 38s & 786.61 & 310.63 & 3m 57s & 1245.25 & 505.75 & 7m 1s \\
    OR Tools+TSP & 416.62 & 185.66 & 1m 39s & \underline{735.52} & 271.47 & 3m 55s & 1178.07 & 426.11 & 7m 19s \\
    \midrule
    
    JOTP & 528.69 & 205.28 & 0.95s & 812.89 & 270.88 & 2.01s & 1001.95 & 332.70 & 3.64s \\
    SABS&465.05&192.74&1m 50s&792.42&268.37&4m 23s&978.12&329.74&8m 15s\\
    \midrule
    
    \sysname & \textbf{406.79} & \textbf{152.03} & 5.06s & \textbf{653.41} & \textbf{217.02} & 8.51s & \textbf{828.44} & \textbf{275.68} & 16.34s \\
    Improvement & 1.8\%& 17.6\%&-&11.2\%&16.1\%&-&9.6\%&12.5\%&-\\
    \bottomrule[1.1pt]
    
    \end{tabular}
  \label{tab:main_1}
\end{table*}

\section{Experiments}
\subsection{Experiment Settings}

\textit{Datasets.} We evaluate our method across two distinct scenarios, yielding a total of six datasets. 
The scenarios include a Real-World environment sourced from a Geekplus warehouse (spanning 31 days of historical operations on a $40\times 72$ grid map) and a Synthetic environment ($100\times 80$ grid map). 
For each scenario, we construct three datasets of varying scales: Small, Medium, and Large. 
These datasets range in problem size, specifically regarding the number of robots ($N_r\in\{15,20,25\}$) and total orders ($N_o\in\{200,500,1000\}$). 
Detailed statistical parameters are provided in Appendix~\ref{app:dataset}.

\textit{Baselines.} We compare \sysname against comprehensive baselines categorized into Phased Methods and Joint Methods. The phased methods combine three order allocation heuristics---Shortest Queue First (SQF)~\cite{sqf}, Work Load Balance (WLB)~\cite{wlb}, and an OR-Tools solver---with four robot scheduling strategies: Nearest\cite{nearest}, Earliest\cite{earliest}, TSP\cite{tsp}, and PSMDRL\cite{psmdrl}. For joint optimization, we include state-of-the-art algorithms such as JOTP\cite{jotp} and SABS\cite{yang2021joint}. The implementation details are provided in Appendix~\ref{app:baseline}.

\textit{Evaluation Metrics.} Performance is assessed using three key metrics: (1) Makespan (Obj), defined as the maximum completion time across all robots, serving as a proxy for overall system throughput; (2) Average Order Completion Time (CompT), which measures the average duration from order arrival to fulfillment, reflecting service responsiveness; and (3) Computation Time (Time), evaluating the computational efficiency of each method per instance.

For reproduction and evaluation of all methods, we built a simulation environment in which the experiments were conducted. More details of the experiments are provided in Appendix~\ref{app:exp}.

\subsection{Main Results}
As presented in Table~\ref{tab:main_1}, our proposed method explicitly dominates all baselines across both synthetic and real-world datasets, achieving the optimal Makespan and Average Order Completion Time. Specifically, compared to the strongest baseline in real-world scenarios, our framework reduces the makespan by \textbf{7.5\%} and the order completion time by \textbf{15.4\%}. 
\sysname\ introduces a soft allocation mechanism to achieve joint optimization. Furthermore, the Event-Driven MDP modeling ensures real-time decision-making capabilities, thereby yielding superior performance.

\textit{Comparison with Decoupled Methods.}
Decoupled methods generally maintain high computational efficiency by decomposing the problem. However, their inability to coordinate order allocation and robot scheduling leads to suboptimal resource utilization. While they provide a baseline level of responsiveness, they consistently trail \sysname in global efficiency metrics, confirming that decoupling decision phases inherently limits the system performance.

\textit{Comparison with Joint Methods.}
Joint methods do not exhibit a significant advantage over phased methods. We attribute this to their reliance on a Rolling-Horizon mechanism to adapt static optimization models to dynamic environments. First, the high computational overhead of solving static joint models causes significant delays, rendering decisions outdated by the time they are executed; Second, optimizing discrete static time windows fails to capture the continuous, long-term evolution of the system.

\textit{Computational Efficiency.} 
Regarding inference speed, while our deep learning-based approach incurs a marginal overhead compared to simple heuristics, it is significantly faster than the SABS and OR-Tools.  In practical deployment, our model achieves a single-phase decision latency in \textbf{sub-100ms}, fully satisfying the real-time responsiveness requirements.

\subsection{Ablation Studies}
\begin{table}[t]
  \centering
  \caption{Results of ablation studies.}
  \label{tab:ablation}
  \renewcommand{\arraystretch}{0.9}   
  \resizebox{\columnwidth}{!}{       
    \begin{tabular}{l|cc|cc|cc}
    \toprule[1.1pt]
    \textbf{Dataset} & \multicolumn{2}{c|}{\textbf{Synth-Small}} & \multicolumn{2}{c|}{\textbf{Synth-Medium}} & \multicolumn{2}{c}{\textbf{Synth-Large}} \\
    \midrule
    \textbf{Metrics} & Obj ($\downarrow$) & CompT ($\downarrow$) & Obj ($\downarrow$) & CompT ($\downarrow$) & Obj ($\downarrow$) & CompT ($\downarrow$) \\
    \midrule
    all & \textbf{811.01} & \textbf{243.25} & \textbf{1387.82} & \textbf{289.94} & \textbf{2160.47} & \textbf{362.08} \\
    w/o Soft & 1184.12 & 463.72 & 2047.31 & 649.60 & 2895.31 & 750.07 \\
    w/o HGT & 826.16 & 275.40 & 1508.50 & 370.47 & 2280.69 & 373.68 \\
    RS-Sum & 895.44 & 269.81 & 1440.56 & 316.95 & 2347.62 & 391.16 \\
    RS-Max & 877.30 & 265.84 & 1447.04 & 320.25 & 2271.81 & 376.91 \\
    w/o Bias & 882.03 & 271.29 & 1455.09 & 316.95 & 2403.80 & 390.21 \\
    Only Bias & 980.74 & 291.81 & 1614.72 & 351.31 & 2437.38 & 403.98 \\
    \midrule
    \midrule
    \textbf{Dataset} & \multicolumn{2}{c|}{\textbf{Real-Small}} & \multicolumn{2}{c|}{\textbf{Real-Medium}} & \multicolumn{2}{c}{\textbf{Real-Large}} \\
    \midrule
    \textbf{Metrics} & Obj ($\downarrow$) & CompT ($\downarrow$) & Obj ($\downarrow$) & CompT ($\downarrow$) & Obj ($\downarrow$) & CompT ($\downarrow$) \\
    \midrule
    all & \textbf{406.79} & \textbf{152.03} & \textbf{653.41} & \textbf{217.02} & \textbf{828.44} & \textbf{275.68} \\
    w/o Soft & 509.98 & 207.37 & 774.18 & 269.16 & 940.42 & 325.02 \\
    w/o HGT & 433.67 & 178.13 & 725.52 & 260.23 & 1003.87 & 395.11 \\
    RS-Sum & 420.38 & 166.89 & 683.73 & 221.10 & 876.51 & 289.67 \\
    RS-Max & 427.70 & 168.31 & 665.98 & 226.78 & 850.46 & 295.22 \\
    w/o Bias & 424.15 & 167.29 & 689.00 & 237.36 & 927.67 & 360.04 \\
    Only Bias & 446.27 & 174.91 & 696.46 & 229.84 & 886.97 & 302.02 \\
    \bottomrule[1.1pt]
    \end{tabular}
  }
\end{table}

The settings of the ablation experiments are as follows:

\begin{itemize}[leftmargin=*]
	\item \textbf{w/o Soft Allocation}: Replace soft allocation with random allocation.
	\item \textbf{w/o HGT}: The Heterogeneous Graph Transformer is replaced with a Transformer~\cite{attention}, where the relationships among heterogeneous entities are modeled in a uniform manner.
	\item \textbf{RS-Sum:} utilizes the cumulative potential $\Phi(s)=\sum_{r \in \mathcal{R}}T_r$.
    \item \textbf{RS-Max:} utilizes the bottleneck potential $\Phi(s)=\max_{r \in \mathcal{R}}T_r$.
    \item \textbf{w/o Bias:} remove the phase-specific bias.
    \item \textbf{Only Bias:} select action with the highest bias value.
\end{itemize}

The experimental results are presented in table~\ref{tab:ablation}. The exclusion of soft allocation exerts the most significant impact on performance. This is attributed to the model's inability to defer decision-making, thereby failing to account for the dynamic variations within the system. Replacing the HGT with a standard Transformer also results in performance degradation. This stems from the standard Transformer's incapacity to model distinct relationships among heterogeneous entities. The reward shaping also makes a contribution to performance. Using phase-bias alone is slightly worse than the decoder-only method except for dataset Real-Large. This is because this phase-knowledge are raw features in the decision model. Adding phase-bias directly only guides the effective exploration in the early stages of reinforcement learning training, while complex global decision information still needs to rely on deep neural networks for modeling.
\subsection{Sensitivity Analysis}
\begin{figure}[t] 
    \centering
    \begin{subfigure}[b]{0.49\linewidth}
        \centering
        \includegraphics[width=\linewidth]{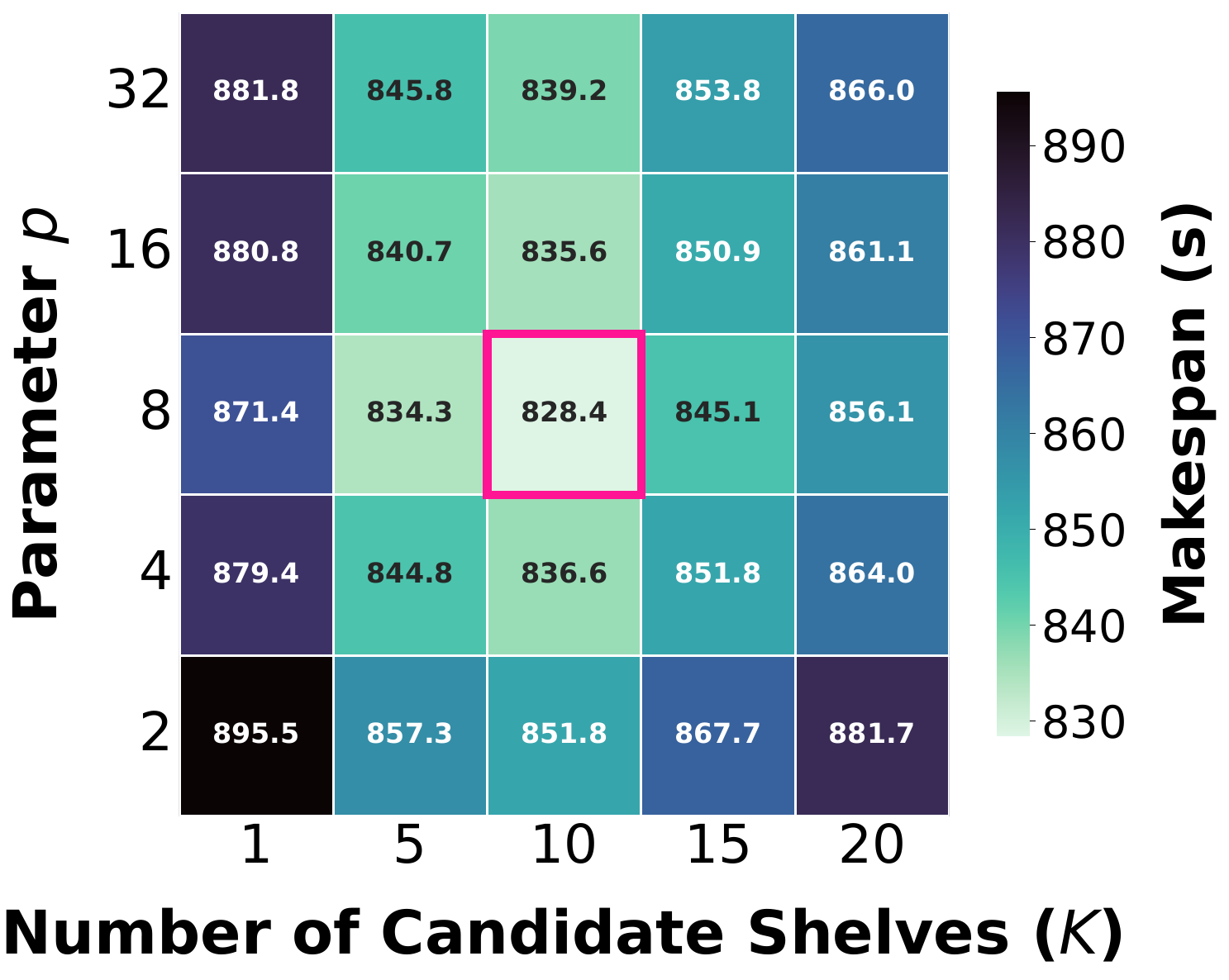}
        \caption{Real-Large}
        \label{fig:a}
    \end{subfigure}
    \hfill 
    \begin{subfigure}[b]{0.49\linewidth}
        \centering
        \includegraphics[width=\linewidth]{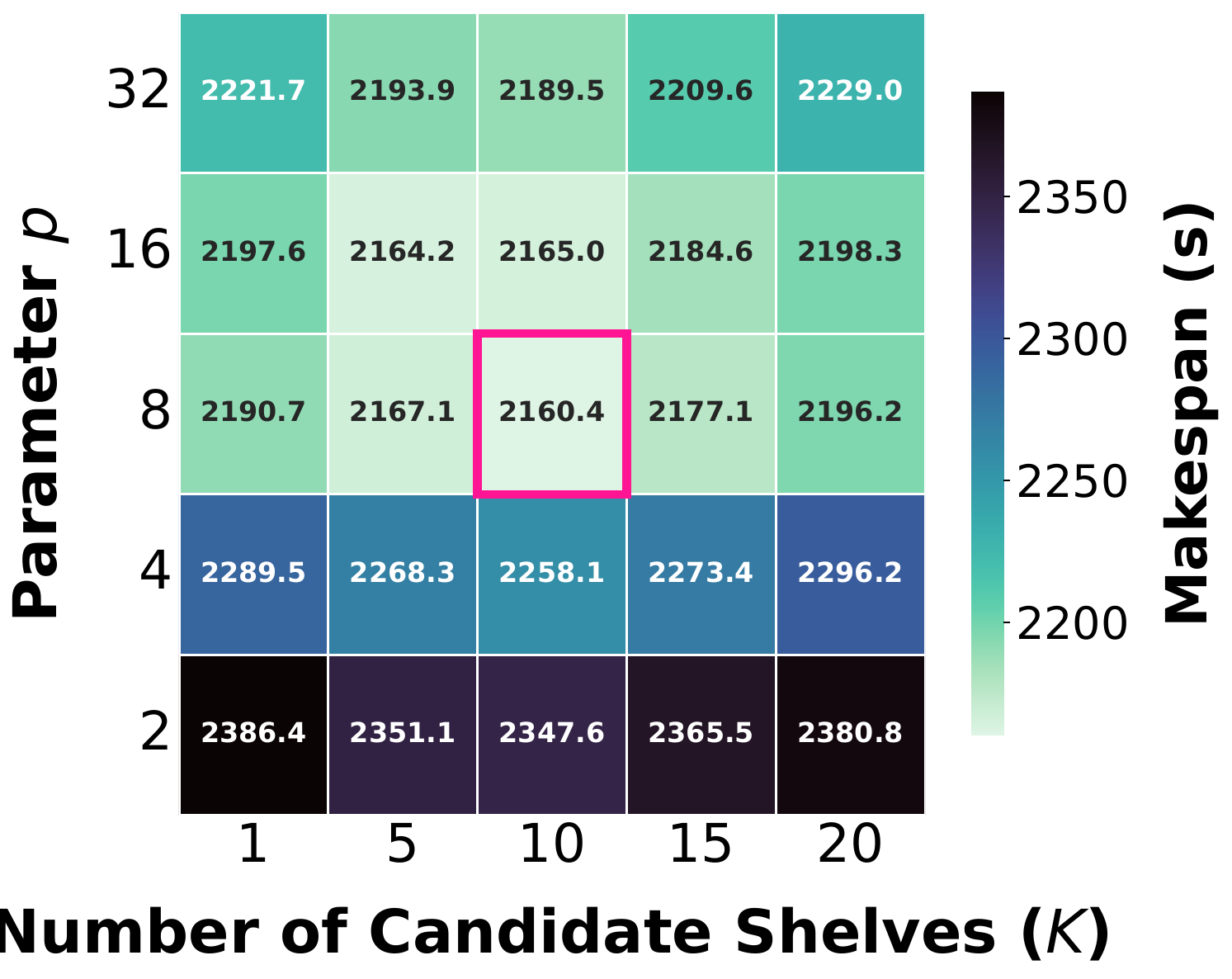}
        \caption{Synth-Large}
        \label{fig:b}
    \end{subfigure}
    
    \caption{Sensitivity analysis of $K$ and $p$ in Large datasets.}
    \label{fig:sensitivity_all}
\end{figure}

    
We analyzed the candidate shelf size $K\in\{1,5,10,15,20\}$ and reward shaping parameter $p\in\{2,4,8,16,32\}$ on the Large datasets. The experimental results are presented in Fig.~\ref{fig:sensitivity_all}.

\textit{Impact of Candidate Shelf Size ($K$):} The Makespan follows a U-shaped trend, achieving optimality at $K=10$. A small $K$ limits the global perspective, degenerating into a greedy local search. Conversely, an excessive $K$ ($K>10$) introduces low-relevance noise into the HGT, increasing state complexity and interfering with critical feature extraction.

\textit{Impact of Reward Shaping Parameter ($p$):} Lower values prioritize cumulative time rather than the makespan, failing to penalize system bottlenecks. Higher values ($p=32$), while theoretically closer to the $L_\infty$ norm, create steep optimization landscapes that cause training instability and convergence to local optima.

\section{Real-World Deployment}
To further evaluate the actual performance of \sysname, we deployed it in a typical e-commerce RMFS environment. The warehouse covers a total area of 5,158 $m^2$ and is equipped with 861 shelves, 16 workstations, and seeding walls comprising 40 slots. The entire cluster consists of 198 robots, operating collaboratively under high-concurrency scenarios with a daily order volume exceeding 13,000.
\subsection{Digital Twin Platform}
\begin{figure}[t]
	\centering
	\includegraphics[width=0.9\linewidth]{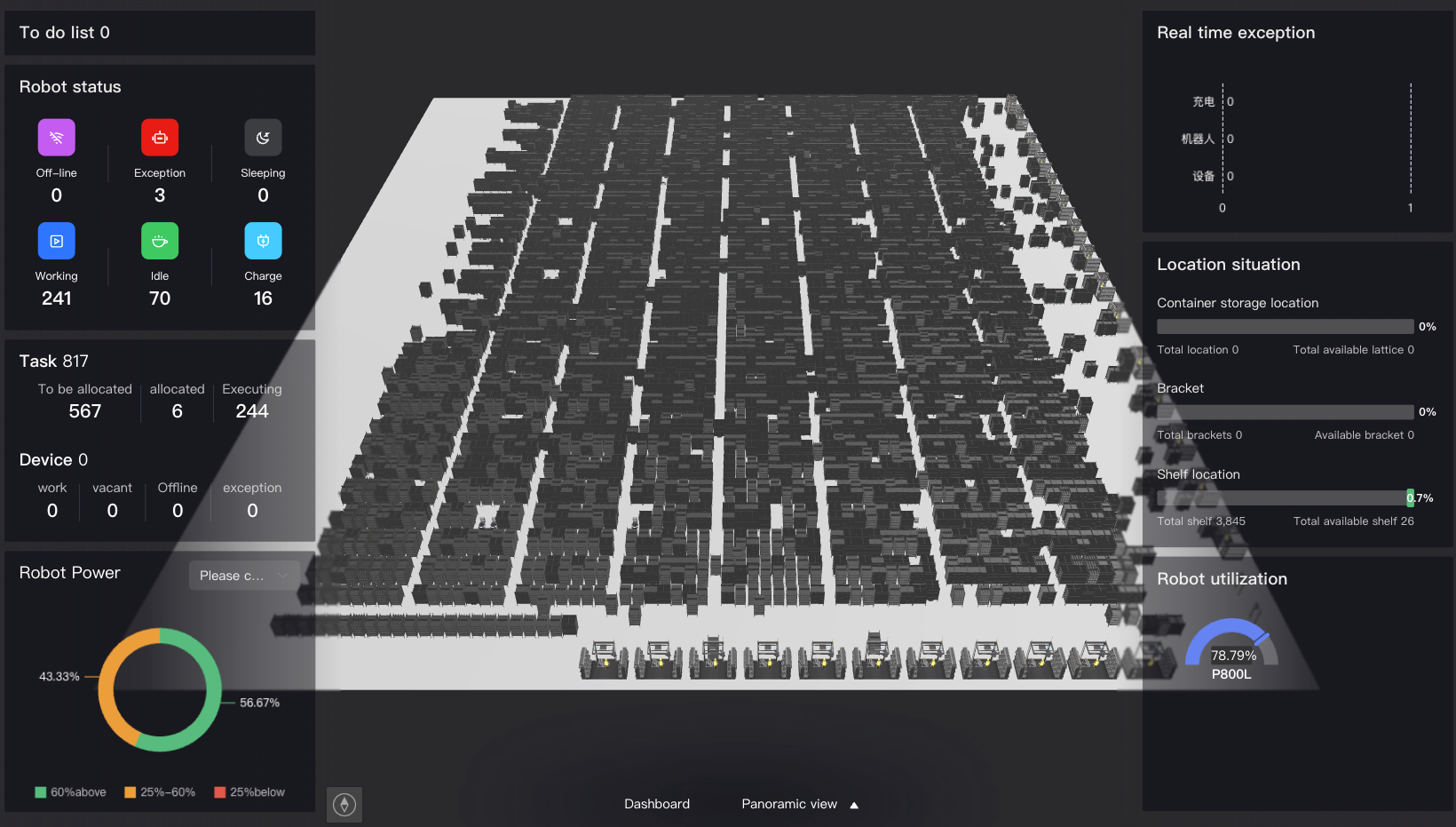}
	\caption{Digital Twin Platform.}
	\label{fig:sim}
\end{figure}

To ensure the robustness of the policy during deployment, we used a high-fidelity digital twin platform based on the actual production environment, as shown in Fig. \ref{fig:sim}. This platform directly integrates the control kernel and physical kinematic models of the production-grade Robot Management System, thereby ensuring that the policy derived from simulation training can be seamlessly transferred to the online environment.

\subsection{Sim-to-Real Adaptation}

\textit{Episode Alignment.} We segment training episodes according to worker rest schedules (approximately 4 to 6 hours). During these breaks, the order backlog is naturally cleared, which effectively disentangles state dependencies, ensuring that each experimental cycle starts from a fresh state without mutual interference.

\textit{Path Planning.} \sysname\ framework provides high-level destination selection, while the lower-level planner plans an efficient, collision-free path from the current location to the target location in millisecond-level. This ensures a closed loop of efficient decision-making and safe execution.

\textit{Entities Pruning for Inference Scaling.} To address latency in real-world scenarios, we introduce an Entities Pruning Mechanism. Specifically, for the two most numerous entities robots and storage locations, we perform the following processing: \textbf{1) Robots:} Only encode the $K_1$ robots closest to the decision-making robot; \textbf{2) Storage Locations:} Only encode the storage locations of the Top-$K_2$ shelves ranked by $\bm{h}^{S}$ and empty storage locations. This optimization reduces the number of encoded entities and caps inference latency at 100 ms.

\subsection{Real-World Evaluation}
\label{sec:real_world_exp}

To validate the robustness of \sysname in a dynamic industrial setting, we deployed the framework into the production environment for a consecutive 7-day field test. 

\textit{Testing via Digital Twin.} 
Conducting fair comparisons in a live warehouse is challenging due to the high variance in order volume and structure across different time windows. To overcome this, we implemented a testing strategy utilizing the digital twin platform. 
While \sysname governed the physical robots in the real world, we simultaneously executed all baseline methods within the digital twin, feeding them the \textbf{identical mirrored real-time order stream}. 

\textit{Baseline Selection.} 
We benchmarked \sysname against the fast, competitive methods from our simulation experiments: OR Tools+TSP, JOTP, WLB+Nearest, and WLB+TSP. Additionally, we included the \textbf{Prod Heuristic}, a phased heuristic algorithm currently applied in system, to represent the current industrial standard.

\textit{Analysis of Comprehensive Performance Metrics.}
As detailed in Table~\ref{tab:real_exp}, we evaluated the system across four core dimensions:

\textbf{(1) Workstation Throughput (Efficiency):} defined as the number of shelves processed per workstation per hour. 
    \sysname achieved a \textbf{3.85\% increase} over the prod heuristic. This improvement indicates that \sysname  effectively balances the workload across workstations and accelerates transportation.
    
\textbf{(2) Average Order Completion Time (Service Level):} defined as the latency from order arrival to picking completion. 
    Our method reduced this metric by \textbf{55 seconds} on average. 
    
\textbf{(3) Shelf Hit Rate (Optimization Quality):} defined as the average number of required items picked per shelf entry. 
    We observed a significant \textbf{6\% increase} in hit rate. This indicates that our framework successfully learned an order batching strategy based on the soft allocation mechanism.
    
\textbf{(4) Average Robot Travel Distance (Sustainability):} defined as the average cumulative distance traveled by a robot to fulfill the same batch of orders. 
    The average travel distance decreased by \textbf{2.81\%}. This is a direct consequence of the improved Shelf Hit Rate: fewer trips are required to fulfill the same demand. 
The marginal performance gap between the digital twin platform and the real-world deployment also confirms the effectiveness of the evaluation. Due to the strict time constraints, the performance of this OR Tools degrades significantly.

\begin{table}[t]
  \centering
  \caption{Performance on real-world deployment.}
  \label{tab:real_exp}  
  \renewcommand{\arraystretch}{0.9} 
  \resizebox{1\columnwidth}{!}{
    \begin{tabular}{l|cccc}
    \toprule[1.1pt]
    \textbf{Metrics} & \textbf{Throughput} ($\uparrow$) & \textbf{CompT} ($\downarrow$) & \textbf{Hit Rate} ($\uparrow$) & \textbf{Distance} ($\downarrow$) \\
    \midrule
    OR Tools+TSP & 28.61 & 2104.77 & 1.20 &  35705.47 \\
    JOTP & 37.27 & 1862.41 & 1.21 & 32105.81 \\
    WLB+Nearest & 38.34 & 1765.84 & 1.24 & 31250.54 \\
    WLB+TSP & 39.02 & 1753.38 & 1.28 &  30420.23 \\
    \midrule
    Prod Heuristic & 41.95 & 1624.46 & 1.32 & 29451.66 \\
    \midrule
    \sysname & \textbf{43.56} & \textbf{1569.52} & \textbf{1.40} & \textbf{28623.80} \\
    \sysname (Real) & 43.41 & 1574.57 & 1.41 & 29035.91 \\
    \bottomrule[1.1pt]
    \end{tabular}
  }
\end{table}



\section{Related Works}
Our work relates to phased methods, including order allocation and robot scheduling, as well as joint methods. Details are as follows:

\textit{Order Allocation Methods.} The order allocation problem is frequently formulated as a matching problem involving orders, shelves, and workstations. Some approaches model the global matching process as an Integer Programming problem~\cite{order1,order5}, while others~\cite{order2,order3,order4} employ algorithms related to bipartite graph matching to derive solutions.

\textit{Robot Scheduling Methods.}
Static robot scheduling problems are frequently formulated as Vehicle Routing Problems (VRP)~\cite{rob1,gharehgozli2020robot,vrp1,vrp2} or Integer Programming problems~\cite{zhuang2022rack,rob3,rob4}. Conversely, dynamic robot scheduling problems are typically solved using reinforcement learning techniques. Early studies~\cite{papoudakis2020benchmarking,christianos2020shared,krnjaic2024scalable} adopted a distributed decision-making paradigm, which necessitated determining the step-wise movement of every robot. Subsequent research~\cite{psmdrl,zhou2024scalable} decided the destination only.

\textit{Joint Methods.}
Existing joint decision-making methodologies primarily concentrate on static problems. These approaches typically formulate the problem mathematically and subsequently employ meta-heuristic algorithms~\cite{yang2021joint} or Mixed-Integer Programming (MIP)~\cite{teck2022bi,zhao2025learning} to derive solutions. 

Compared with these methods, \sysname\ introduces a soft allocation mechanism and an event-driven decision-making mechanism to enable real-time joint decision-making.
\section{Conclusions}
This paper presents \sysname, a real-time system that jointly optimizes order allocation and robot scheduling for Robotic Mobile Fulfillment Systems (RMFS). 
By leveraging a soft allocation mechanism and an event-driven decision strategy, \sysname significantly enhances system throughput while ensuring real-time responsiveness. 
Extensive experiments on both synthetic and real-world datasets, along with sim-to-real deployment, demonstrate that \sysname effectively bridges the gap between real-time responsiveness and global optimization, showing promising application prospects. 
Currently, the framework focuses on high-level scheduling optimization and has not yet integrated physical-level controls such as path planning and collision avoidance. 
In future work, we aim to integrate physical execution  to realize a more comprehensive, end-to-end framework.

\balance
\bibliographystyle{ACM-Reference-Format}
\bibliography{sample-base}

\appendix

\newpage
\section{Dataset Details}
\label{app:dataset}
\begin{table}[htbp]
    \centering
    \small
    
    \caption{Summary of Dataset Parameters}
    \label{tab:combined_setup}

    \begin{subtable}[t]{0.42\linewidth} 
        \centering
        \caption{Warehouse Settings}
        \begin{tabular}{l|c|c} 
            \toprule
             & Real & Synth \\ 
            \midrule
            $N_s$ & 861 & 1600 \\
            $N_w$ & 16 & 23 \\
            Scale & $40{\times}72$ & $100{\times}80$ \\ 
            $\text{C}_{\text{shelf}}$ & 5 & 10 \\
            $\text{C}_{\text{item}}$ & 2 & 4 \\
            \bottomrule
        \end{tabular}
        \label{tab:scenario}
    \end{subtable}
    \hfill 
    \begin{subtable}[t]{0.56\linewidth}
        \centering
        \caption{Problem Scales}
        \begin{tabular}{l|c|c|c}
            \toprule
             & Small & Medium & Large \\ 
            \midrule
            $N_r$ & 15 & 20 & 25 \\
            $N_o$ & 200 & 500 & 1000 \\
            \bottomrule
        \end{tabular}
        \label{tab:scale}
    \end{subtable}
\end{table}

To comprehensively evaluate \sysname, we conducted experiments on six datasets encompassing both real-world and synthetic scenarios across three distinct scales. The static characteristics of these scenarios are summarized in Table~\ref{tab:scenario}, with the scale specifications detailed in Table~\ref{tab:scale}. \textbf{The items processing time is calculated by: ${num_{item}}\times C_{\text{item}}+C_{\text{shelf}}$}. The specific data sources and processing procedures for these datasets are described as follows:

\begin{itemize}[leftmargin=*]
\item \textbf{Real Dataset}: Sourced from a real-world Geekplus warehouse, comprising actual layouts, inventory snapshots, and 31 days of historical order data. We chronologically partitioned the data into training (21 days), validation (5 days), and test (5 days) sets. To ensure operational intensity, sub-problems were sampled based on high order density criteria. Both the validation and test sets contain 100 instances.
 \item \textbf{Synthetic Dataset}: To simulate the wave release characteristics inherent in warehousing systems, order arrival times are not distributed uniformly; instead, they follow a wave distribution incorporating random perturbations. We assume a total of $N_{total}$ orders, with the number of orders per wave set to $S_{wave}$ (defaulting to 50) and the wave interval set to $T_{wave}$ (defaulting to 60 seconds). For the $i$-th order, the arrival time $t_i$ is generated according to the following formula:
 \begin{equation}
		t_i = k \cdot T_{wave} + \epsilon, \quad k \sim U\{0, N_{waves}-1\}, \quad \epsilon \sim U\{-3, 3\}
	\end{equation}
	where $N_{waves} = \lceil N_{total} / S_{wave} \rceil$ represents the total number of waves, $k$ denotes the randomly sampled wave index, and $\epsilon$ represents uniform random noise within the range $[-3, 3]$ seconds, intended to simulate minute temporal discrepancies in order release times within a single wave.
	
	Consistent with the characteristics of actual warehousing data, both the number of distinct item types contained in an order (Number of Order Lines, $L_i$) and the specific quantity required for each item (Quantity per item, $Q_{ij}$) follow a Truncated Pareto Distribution. This distribution is selected to capture the long-tail characteristic typically described as ``few large, many small.''
	Specifically, let $\text{Pareto}(\alpha)$ denote a random variable following a Pareto distribution with a shape parameter $\alpha$ (set to $\alpha=2$ in the experiments). The generated values are calculated as follows:
	\begin{equation}
		X = \min \left( \lfloor \text{Pareto}(\alpha) + 1 \rfloor, X_{max} \right)
	\end{equation}
	For the number of order lines $L_i$, the truncation upper limit $X_{max}$ is set to $L_{max}$ (defaulting to 4); for the quantity per item $Q_{ij}$, the truncation limit is set to $Q_{max}$ (defaulting to 4). The specific item IDs are sampled uniformly at random from the set of items with non-zero current inventory.
\end{itemize}


\section{Default Order Allocation Strategy}
\label{app:hard_allocate}
\begin{algorithm}
	\caption{Default Order Allocation}
	\label{alg:hard}
	\begin{algorithmic}[1]
		\Require Arriving order $o$; 
            Set of workstations $\mathcal{W}$; 
            Set of shelves $\mathcal{S}$; 
		\Ensure The workstation allocated to order $w_o$; The shelves set allocated to order $\mathcal{S}_o$
		\State Select allocated workstation $w_o$ based on rules
		\State $\mathcal{S}_o\gets \emptyset$
		\While{$\bm{d}_o \neq \bm{0}$}
		\For {each shelf $s \in \mathcal{S}$}
		\State Calculate matching degree $\bm{V}^{(o)}_{s, w}$ via Eq.~\eqref{eq:matching_degree}
		\EndFor
		\State Select the shelf $s'$ with the maximum matching degree
		\State $\mathcal{S}_o\gets \mathcal{S}_o \cup\{s'\}$
		\State $\bm{a}\gets \min(\bm{d}_{o},\bm{q}_{s'})$
		\State $\bm{d}_{o}\gets\bm{d}_{o}-\bm{a}$
		\State $\bm{q}_{s'}\gets\bm{q}_{s'}-\bm{a}$
		\EndWhile
		\State \textbf{return} $w_o$, $\mathcal{S}_o$
	\end{algorithmic}
\end{algorithm}

Upon the pick-up of a shelf by a robot, the soft order allocation method allocates all soft-allocated orders to that shelf; however, the items on the shelf may be insufficient. To address this deficiency, it is necessary to allocate the order to additional shelves to ultimately satisfy the order requirements. The specific methodology is described in Algorithm~\ref{alg:hard}. This method is also employed in several baselines.

\section{Baseline Details}
\label{app:baseline}
We selected two categories of methodologies: phased decision-making methods and joint decision-making methods. For the phased methods, we selected three order allocation methods and three robot scheduling methods, respectively, and evaluated the performance of all nine resulting combinations.

\subsection{Order Allocation Methods}
\begin{itemize}[leftmargin=*]
    \item \textbf{Shortest Queue First (SQF) ~\cite{sqf}}: This method allocates the order to the workstation with the shortest expected queue time and subsequently selects shelves based on Algorithm~\ref{alg:hard}.
    \item \textbf{Work Load Balancing (WLB) ~\cite{wlb}}: 
    This method allocates the order to the workstation with the minimum workload and subsequently selects shelves based Algorithm~\ref{alg:hard}.
    \item \textbf{OR Tools}: This method model the dynamic order allocation as a static problem via a rolling-horizon approach and solve it using the Google OR-Tools CP-SAT solver.
\end{itemize}

\subsection{Robot Scheduling Methods}
\begin{itemize}[leftmargin=*]
    \item \textbf{Nearest Neighbor~\cite{nearest}}: This method selects the nearest valid target.

    \item \textbf{Earliest Arrival Order~\cite{earliest}}: This method selects the earliest arriving order for completion. Furthermore, to adapt it to our three-phase decision problem, its return strategy is set to the nearest neighbor approach.
    
    \item \textbf{TSP~\cite{tsp}}: 
    This method models the robot scheduling process as a Traveling Salesman Problem.

    \item \textbf{PSMDRL~\cite{psmdrl}}: A reinforcement learning method employs a Transformer architecture for decision.
\end{itemize}

\subsection{Joint Optimization Methods}

\begin{itemize}[leftmargin=*]
    \item \textbf{JOTP~\cite{jotp}}: 
    A joint optimization method combining the Kuhn-Munkres algorithm~\cite{km} and reinforcement learning.
    \item \textbf{SABS~\cite{yang2021joint}}:
    A hybrid heuristic algorithm combining simulated annealing and beam search for the joint optimization.
\end{itemize}
\section{Evaluation Metrics}
To ensure a comprehensive assessment of system performance, we utilize three key metrics:
\begin{itemize}[leftmargin=*]
\item \textbf{Makespan (Obj):} Defined as the maximum completion time among all robots. This metric indicates the total time required to clear the current batch of orders, serving as a direct proxy for the system's overall throughput and efficiency.
\item \textbf{Average Order Completion Time (CompT):} Defined as the average duration between the arrival of an order and its final fulfillment. This metric reflects the system's responsiveness and the quality of service provided to individual orders.
\item \textbf{Time:} The average computation time per problem instance.
\end{itemize}

\section{Entity Encoding and Value Function Features Details}
\label{app:encode}
The continuous features to be modeled for robot $r$, storage location $l$, and workstation $w$ are as follows:

\begin{equation}
	\bm{f}_r^{raw}=x_r\Vert y_r\Vert{task}_r\Vert {h}_r\Vert {soft}_{r}.
\end{equation}

\begin{equation}
	\bm{f}_l^{raw}=x_l\Vert y_l\Vert dist_l\Vert {task}_l\Vert {h}_l\Vert {soft}_l.
\end{equation}

\begin{equation}
	\bm{f}_w^{raw}=x_w\Vert y_w\Vert dist_w\Vert {task}_w\Vert {h}_w\Vert{u}_w\Vert{cost}_w.
\end{equation}

Here, $x, y$ represent the entity coordinates; $task$ denotes the number of tasks (when processing orders in $\mathcal{O}_{rem}$ that cannot be resolved by soft allocation, additional shelves need to be allocated and moved to workstations; each move is called a task) assigned to the entity (or its carried shelf for robot, its shelf for storage location, hereinafter the same); $dist$ denotes the distance to the decision-making robot; $h$ represents the soft allocation heat of the entity; and $soft$ indicates the size of soft orders set for the entity. $u$ represents the workload of the workstation, and $cost$ indicates the potential increase in total queuing time resulting from proceeding to that workstation. 

Robots and storage locations exhibit high state dynamics. Furthermore, there are significant differences in decision-making logic across entities in different status. To model this discrete state information, we sum the learnable state embedding vectors with the continuous feature vectors to obtain the final representation for each entity:
\begin{equation}
	\bm{h}^0_{r}={\text{MLP}_r(\bm{f}_r^{raw})+\bm{e}_{r}({status}_r)}.
\end{equation}

\begin{equation}
	\bm{h}^0_{l}={\text{MLP}_l(\bm{f}_l^{raw})+\bm{e}_{l}({status}_l)}.
\end{equation}

\begin{equation}
	\bm{h}^0_{w}={\text{MLP}_w(\bm{f}_w^{raw})}.
\end{equation}

$\bm{h}^0_{r}\in \mathbb{R}^D$, $\bm{h}^0_{l}\in \mathbb{R}^D$, and $\bm{h}^0_{w}\in \mathbb{R}^D$ are the dense embedding vectors for robot $r$, storage location $l$, and workstation $w$, respectively. ${status}_r$ denotes the scheduling phase for $r$ and ${status}_l$ denotes the shelf occupancy status for $l$. We set the dimension of the embedding vectors for all entities to a uniform value $D$.

For PPO value function, we use a linear layer that takes a feature vector comprising the number of remaining orders, completed soft orders, and completed tasks as input.

\section{Heterogeneous Graph Transformer}
\label{app:hgt}
HGT employs $L$ layers of heterogeneous attention convolutions. Relation set $\mathcal{T}$ is defined as $\{\tau_{rw}, \tau_{wr}, \tau_{rl}, \tau_{lr}, \tau_{wl}, \tau_{lw}\}$. Each layer consists of a multi-relational aggregation operator, $\mathrm{HeteroConv}$, which applies a variant of GAT\cite{gatv2} with edge features for each relation $\tau\in\mathcal{T}$:
\begin{equation}
	\bm{h}_{v}^{(\ell+1)}=\sum_{\tau\in\mathcal{T}}\mathrm{GAT}_{\tau}\!\left(\bm{h}^{(\ell)},\,\bm{D}\right),\quad \ell=0,\dots,L-1,
\end{equation}

where $\bm{D}$ represents the distance matrix between entities. The computation of the GAT operator is defined as follows:

\begin{equation}
	\bm{h}'_i = \sum_{j \in \mathcal{N}(i) \cup \{i\}} \alpha_{i,j} \bm{W}_t \bm{h}_j,
\end{equation}

\begin{equation}
	\alpha_{i,j} = \frac{\exp \left( \bm{a}^\top \text{LeakyReLU} \left( \bm{W}_s \bm{h}_i + \bm{W}_t \bm{h}_j + \bm{W}_e \bm{D}_{i,j} \right) \right)}{\sum_{k \in \mathcal{N}(i) \cup \{i\}} \exp \left( \bm{a}^\top \text{LeakyReLU} \left( \bm{W}_s \bm{h}_i + \bm{W}_t \bm{h}_k + \bm{W}_e \bm{D}_{i,k} \right) \right)}.
\end{equation}

The aggregation method across relations is summation. To stabilize training and preserve original representations, residual connections and layer normalization are applied to each node type at every layer:
\begin{equation}
	\tilde{\bm{h}}_{\tau}^{(\ell+1)}=\mathrm{LayerNorm}\!\left(\bm{h}_{\tau}^{(\ell+1)}+\bm{h}_{\tau}^{(\ell)}\right),\ \forall \tau\in\mathcal{T}.
\end{equation}

\section{Detailed Experimental Settings}
\label{app:exp}
In this section, we detail the experimental setup. All experiments were conducted using Python 3.9.23 and PyTorch 2.4.1.

Regarding the reinforcement learning algorithm, we employed Proximal Policy Optimization (PPO) as the core training algorithm, utilizing the following parameter configuration: The number of local parallel environments is set to $8$, with each environment collecting $128$ time steps of data prior to each policy update. Upon the completion of data collection, the policy network undergoes updates for $4$ epochs. To enhance computational efficiency, the data is further partitioned into $4$ minibatches for gradient descent. The PPO clip coefficient $\epsilon$ is set to $0.1$ to constrain the magnitude of policy updates and prevent drastic policy oscillations. Concurrently, the entropy coefficient is set to $0.01$ to encourage exploration, and the value function coefficient is set to $0.1$. To mitigate gradient explosion, the maximum gradient norm is limited to $0.5$. Furthermore, we established a target KL divergence threshold of $0.01$; should the approximate KL divergence exceed this value, the updates for the current epoch are terminated prematurely. The total number of timesteps for the entire training process is set to $4 \times 10^6$.

Regarding the model architecture, we employed a Heterogeneous Graph Attention Network comprising $4$ layers of heterogeneous attention convolutions. The hidden size $D$ is set to $256$, and the number of attention heads is set to $2$. 

For hyperparameter in \sysname, the size of the candidate set $K$ is set to $10$, the threshold for the number of encoded robots ($K_1$) and storage locations ($K_2$) is set to $50$, the reward shaping hyperparameter $p$ is set to $8$.

During the training process, we monitored the model's performance on the validation set and retained the best-performing model as the final model.

\section{OR Tools Implementation Details}

We employ a rolling-horizon approach to transform the dynamic order allocation problem into a static problem. Specifically, we maintain a dynamic order pool into which incoming orders are accumulated. When the number of orders in the pool reaches a specific threshold or the time elapsed since the previous allocation exceeds a designated time limit, Google OR-Tools is invoked to determine the order allocation scheme for the current pool, which is subsequently cleared. In the experiments, the order quantity threshold is set to $10$, the elapsed time threshold is set to $60$ time units, and the upper limit of solver solution time is set to $10$ seconds for real dataset, $15$ seconds for synthetic dataset. If the threshold is exceeded, the default allocation method will be used. We formulate the order allocation problem as a Constraint Programming (CP) problem, with the objective of minimizing the total shelf transportation distance required to fulfill the batch of orders, subject to inventory constraints and workstation assignment constraints.

\subsection{Sets and Parameters}
\begin{itemize}[leftmargin=*]
    \item $O$: Set of orders in the current batch.
    \item $S$: Set of candidate shelves containing required items.
    \item $W$: Set of available workstations.
    \item $K$: Set of item types.
    \item $R_{o,k}$: The quantity of items $k$ required by order $o$.
    \item $I_{s,k}$: The inventory quantity of items $k$ on shelf $s$.
    \item $D_{s,w}$: The travel distance between the storage location of shelf $s$ and workstation $w$.
\end{itemize}

\subsection{Decision Variables}
The model utilizes the following decision variables:

\begin{itemize}[leftmargin=*]
    \item $y_{o,w} \in \{0, 1\}$: Binary variable. Equal to 1 if order $o$ is assigned to workstation $w$, and 0 otherwise.
    \item $z_{s,w} \in \{0, 1\}$: Binary variable. Equal to 1 if shelf $s$ is transported to workstation $w$, and 0 otherwise.
    \item $x_{o,k,s} \in \mathbb{Z}_{\ge 0}$: Integer variable. Represents the quantity of item $k$ for order $o$ picked from shelf $s$. Note that the workstation dimension is optimized out in this variable to reduce model complexity.
\end{itemize}

\subsection{Objective Function}
The objective is to minimize the total distance traveled by the shelves to the assigned workstations:
\begin{equation}
    \text{Minimize} \quad \sum_{s \in S} \sum_{w \in W} D_{s,w} \cdot z_{s,w}
\end{equation}

\subsection{Constraints}

\paragraph{1. Order Assignment Constraint:}
Each order must be assigned to exactly one workstation to be processed.
\begin{equation}
    \sum_{w \in W} y_{o,w} = 1, \quad \forall o \in O
\end{equation}

\paragraph{2. Demand Satisfaction Constraint:}
For every order and every required item, the total quantity picked from all shelves must equal the order's requirement.
\begin{equation}
    \sum_{s \in S} x_{o,k,s} = R_{o,k}, \quad \forall o \in O, \forall k \in K \text{ where } R_{o,k} > 0
\end{equation}

\paragraph{3. Inventory Capacity Constraint:}
The total quantity of a specific item picked from a shelf by all orders cannot exceed the shelf's available inventory.
\begin{equation}
    \sum_{o \in O} x_{o,k,s} \le I_{s,k}, \quad \forall s \in S, \forall k \in K \text{ where } I_{s,k} > 0
\end{equation}

\paragraph{4. Shelf-Workstation Coupling Constraint:}
This constraint links the picking variable $x$, the order assignment $y$, and the shelf movement $z$. It ensures that if an order $o$ assigned to workstation $w$ picks any item from shelf $s$, then shelf $s$ must visit workstation $w$.
In the CP-SAT model, this is implemented using logical implication: if shelf $s$ does not visit workstation $w$ ($z_{s,w}=0$), and order $o$ is assigned to $w$ ($y_{o,w}=1$), then no items can be picked from $s$ for $o$.
\begin{equation}
    (z_{s,w} = 0) \land (y_{o,w} = 1) \implies x_{o,k,s} = 0, \quad \forall o \in O, s \in S, w \in W, k \in K
\end{equation}
Equivalently, this enforces that a shelf trip is generated ($z_{s,w}=1$) whenever a picking activity occurs at that workstation.

\section{Ablation Study of Phase-Specific Bias}

\begin{table}[t]
  \centering
  \caption{Results of bias ablation study.}
  \label{tab:ablation_bias}
  \renewcommand{\arraystretch}{1.1}   
  \resizebox{\columnwidth}{!}{       
    \begin{tabular}{l|cc|cc|cc}
    \toprule[1.1pt]
    \textbf{Dataset} & \multicolumn{2}{c|}{\textbf{Synth-Small}} & \multicolumn{2}{c|}{\textbf{Synth-Medium}} & \multicolumn{2}{c}{\textbf{Synth-Large}} \\
    \midrule
    \textbf{Metrics} & Obj ($\downarrow$) & CompT ($\downarrow$) & Obj ($\downarrow$) & CompT ($\downarrow$) & Obj ($\downarrow$) & CompT ($\downarrow$) \\
    \midrule
    \sysname & \textbf{811.01} & \textbf{243.25} & \textbf{1387.82} & \textbf{289.94} & \textbf{2160.47} & \textbf{362.08} \\
    w/o bias & 882.03 & 265.84 & 1455.09 & 316.95 & 2403.80 & 376.91 \\
    Only bias & 980.74 & 291.81 & 1614.72 & 351.31 & 2437.38 & 403.98 \\
    \midrule
    \midrule
    \textbf{Dataset} & \multicolumn{2}{c|}{\textbf{Real-Small}} & \multicolumn{2}{c|}{\textbf{Real-Medium}} & \multicolumn{2}{c}{\textbf{Real-Large}} \\
    \midrule
    \textbf{Metrics} & Obj ($\downarrow$) & CompT ($\downarrow$) & Obj ($\downarrow$) & CompT ($\downarrow$) & Obj ($\downarrow$) & CompT ($\downarrow$) \\
    \midrule
    \sysname & \textbf{406.79} & \textbf{152.03} & \textbf{653.41} & \textbf{217.02} & \textbf{828.44} & \textbf{275.68} \\
    w/o bias & 427.70 & 168.31 & 689.00 & 237.36 & 927.67 & 360.04 \\
    Only bias & 446.27 & 174.91 & 696.46 & 229.84 & 886.97 & 302.02 \\
    \bottomrule[1.1pt]
    \end{tabular}
  }
\end{table}

To verify the synergistic effect between the decoder model and the phase-bias, we conducted ablation study, and the results are shown in the table~\ref{tab:ablation_bias}.

Results show that using phase-bias alone is significantly worse than the \sysname, and is slightly worse than the decoder-only method except for dataset Real-Large. This is because this phase-knowledge are raw features in the decision model. Adding phase-bias directly only guides the effective exploration in the early stages of reinforcement learning training, while complex global decision information still needs to rely on deep neural networks for modeling.

    
    
    

\section{Analysis of Default Order Allocation Strategy's Impact on Performance}
\label{app:long_tail}
\begin{figure}[t]
	\centering
	\includegraphics[width=0.85\linewidth]{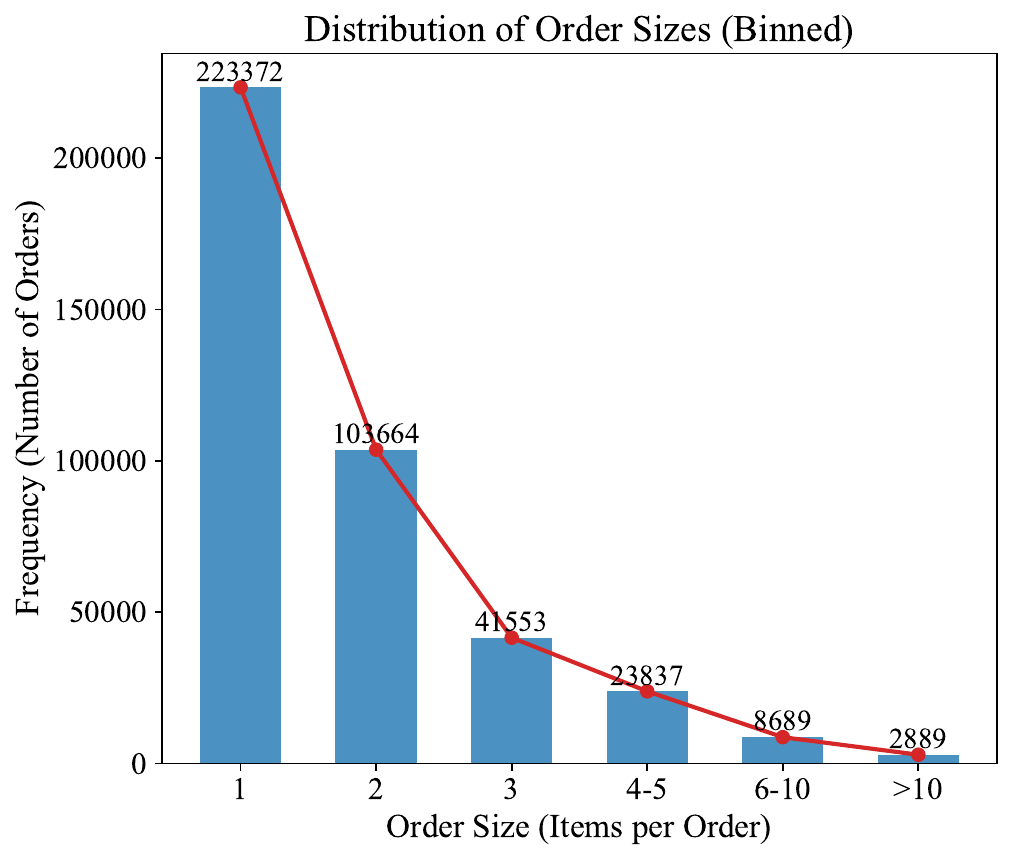}
	\caption{Order Size Distribution.}
	\label{fig:long_tail}
\end{figure}

To assess the impact of the default allocation module on overall system performance, we first analyzed the distribution of order sizes (i.e., the number of items per order), as illustrated in Fig.~\ref{fig:long_tail}. The data exhibits a characteristic long-tail distribution, where approximately 91.2\% of orders contain only 1 to 3 items. This implies that the vast majority of orders can be fulfilled by a single shelf, offering significant potential for order batching. Such a distribution aligns perfectly with our Soft Allocation mechanism, allowing it to process the majority of orders efficiently. Consequently, the default allocation strategy is relegated to handling only the small fraction of complex, large-scale orders, thereby minimizing its impact on overall throughput. To further verify this, we counted the number of orders completed by soft allocation and the default allocation strategy in real-world scenarios 78.3\% of the orders were completed by soft allocation.

\end{document}